\documentclass{article}

\usepackage[final]{neurips_2020}

\usepackage[utf8]{inputenc} 
\usepackage[T1]{fontenc}    
\usepackage{hyperref}       
\usepackage{url}            
\usepackage{booktabs}       
\usepackage{amsfonts}       
\usepackage{nicefrac}       
\usepackage{microtype}      
\usepackage{color,soul}
\usepackage{amsmath}
\usepackage{amsthm}
\usepackage{amstext}
\usepackage{mathrsfs}
\usepackage{amssymb}
\usepackage{amsfonts}
\usepackage{bm}
\usepackage{blkarray}
\usepackage[utf8]{inputenc}
\usepackage{chngcntr}
\usepackage{apptools}
\AtAppendix{\counterwithin{thm}{section}}
\usepackage{amsmath,tikz}
\usetikzlibrary{matrix}
\usepackage[utf8]{inputenc}
\usepackage[english]{babel}
\usepackage{ulem}
\usepackage{sidecap}
\usepackage{subcaption}
\usepackage{hhline}

\makeatletter
\newtheorem*{rep@theorem}{\rep@title}
\newcommand{\newreptheorem}[2]{%
\newenvironment{rep#1}[1]{%
 \def\rep@title{#2 \ref{##1}}%
 \begin{rep@theorem}}%
 {\end{rep@theorem}}}
\makeatother

\newtheorem{thm}{Theorem}
\newreptheorem{thm}{Theorem}
\newtheorem*{thm*}{Theorem}

\newtheorem{prop}[thm]{Proposition}
\newreptheorem{prop}{Proposition}

\theoremstyle{definition}

\newtheorem{definition}[thm]{Definition}
\newtheorem*{definition*}{Definition}

\newtheorem{example}[thm]{Example}
\newtheorem*{example*}{Example}

\newtheorem*{framework*}{Unifying OT Framework}

\newcommand{\fun}[2]{#1\!\left(#2\right)}

\newcommand{\PP}[2]{\fun{P_{#1\!}}{#2}} 

\newcommand{\given}{\vert}

\newcommand{\size}[1]{|#1|}

\newcommand{\dataSpace}{\mathcal D} 
\newcommand{\dataSize}{\size{\dataSpace}}
\newcommand{\concept}{h} 
\newcommand{\conceptSpace}{\mathcal H}
\newcommand{\conceptSize}{\size{\conceptSpace}}
 
\newcommand{\CI}{\mathrm{CI}} 
\newcommand{\CR}{\mathrm{CR}}

\newcommand{\LL}{\PP{L}{\concept\given d}}
\newcommand{\LLpri}{\PP{L_0}{\concept}}
\newcommand{\LLmar}{\PP{L}{d}}

\newcommand{\TT}{\PP{T}{d\given\concept}} 
\newcommand{\TTpri}{\PP{T_0}{d}}
\newcommand{\TTmar}{\PP{T}{\concept}}

\newcommand{\M}{M}

\newcommand{\e}{\epsilon}

\newcommand{\rr}{\mathbf{r}}
\newcommand{\cc}{\mathbf{c}}

\DeclareMathOperator*{\argmax}{arg\,max}
\DeclareMathOperator*{\argmin}{arg\,min}
\DeclareMathOperator*{\arginf}{arg\,inf}

\usepackage{xr}
\makeatletter
\newcommand*{\addFileDependency}[1]{
  \typeout{(#1)}
  \@addtofilelist{#1}
  \IfFileExists{#1}{}{\typeout{No file #1.}}
}
\makeatother

\title{A mathematical theory of cooperative communication}

\author{%
  Pei Wang\\ 
  Rutgers University--Newark\\
  \texttt{peiwang@rutgers.edu}\\
   \And
   Junqi Wang\\
   Rutgers University--Newark \\
  \texttt{junqi.wang@rutgers.edu} \\
   \And
   Pushpi Paranamana\\
   Rutgers University--Newark\\
  \texttt{pushpi.paranamana@nd.edu} \\
   \And
   Patrick Shafto\\
  Rutgers University--Newark\\
  \texttt{patrick.shafto@gmail.com} \\
}

\begin{document}

\maketitle

\begin{abstract}
  Cooperative communication plays a central role in theories of human cognition, language, development, culture, and human-robot interaction. Prior models of cooperative communication are algorithmic in nature and do not shed light on why cooperation may yield effective belief transmission and what limitations may arise due to differences between beliefs of agents. Through a connection to the theory of optimal transport, we establishing a mathematical framework for cooperative communication. We derive prior models as special cases, statistical interpretations of belief transfer plans, and proofs of robustness and instability. Computational simulations support and elaborate our theoretical results, and demonstrate fit to human behavior. The results show that cooperative communication provably enables effective, robust belief transmission which is required to explain feats of human learning and improve human-machine interaction. 
\end{abstract}

  
\section{Introduction}
Cooperative communication is invoked across language, cognitive development, cultural anthropology, and robotics to explain people's ability to effectively transmit information and accumulate knowledge. Theories claim that people have evolved a specialized ecological niche \citep{tomasello1999,boyd2011cultural} and learning mechanisms \citep{csibra2009natural,grice1975logic,sperber1986relevance}, which explain our abilities to learn and accumulate knowledge; however, we lack mathematical theories that would allow us analyze basic properties of cooperative communication between agents. 

Models of belief updating \citep{chater2008probabilistic,tenenbaum2011grow,ghahramani2015probabilistic} and action selection \citep{luce2012individual,sutton1998introduction} have recently been combined into models of cooperative communication in cognitive science \citep{shafto2008teaching,Shafto2014}, cognitive development \citep{Eaves2016c,bonawitz2011double,bridgers2019young}, linguistic pragmatics \citep{goodman2013knowledge}, and robotics \citep{ho2016showing,hadfield2016cooperative,fisac2017pragmatic,smitha2019literal}. These models are algorithms for computing cooperative communication plans using Theory of Mind reasoning. 
However, these models do not formalize the problem mathematically and therefore do not support general conclusions about the nature or limitations of cooperative communication. 


Build upon mathematical and computational analysis, we provide answers to fundamental questions of cooperative communication.
Our contributions are as follows. In Section~\ref{sec:OT}, we interpret cooperative communication as a problem of optimal transport \citep{monge1781memoire,villani2008optimal,peyre2019computational}, derive prior models of cooperative communication as special cases, and derive relationships to rate distortion theory. In particular, we theoretically guarantee the existence of optimal communication plan and algorithmically ensure the achievablility of such plans.
In Section~\ref{sec:analysis}, we mathematically analyze properties of cooperative communication including statistical interpretations, robustness to violations of common ground, and instability under greedy data selection. 
In Section~\ref{sec:exp}, we computationally analyze robustness to common ground violations, sensitivity to greedy selection of data, approximate methods of correcting common ground, and demonstrate fit to human data. 


\section{Cooperative communication as a problem of optimal transport} \label{sec:OT}

Communication is a pair of processes considered between two agents, that we will refer to as a \textit{teacher} and a \textit{learner}, wherein the teacher selects data and the learner draws inferences based on those data. 
Optimal transport provides a mathematical framework for formalizing movement of one distribution to another, and therefore a framework for modeling communication.  
By recasting communication as belief transport we will gain access to mathematical and computational techniques for understanding and analyzing the problem of cooperative communication.

\subsection{Background on Optimal Transport}\label{sec:base_OT}

Optimal Transport has been discovered in many settings and fields \citep{villani2008optimal,kantorovich2006translocation,koopmans1949optimum,dantzig1949programming,brenier1991polar}. 
The general usefulness of optimal transport can be credited to the simplicity of the problem it solves. 
The original formulation, attributable to \cite{monge1781memoire}, involves minimizing the effort required to move a pile of dirt from one shape to another. Where Monge saw dirt, we may see any probability distribution. 


\noindent \textbf{Entropy regularized Optimal Transport.} 
Formally, let $\mathbf{r}=(r_1, \dots, r_n)$ and $\mathbf{c}=(c_1, \dots, c_m)$ be probability vectors of length $n$ and $m$ respectively. 
A joint distribution matrix $P=(P_{ij})$ of dimension $n\times m$ is called a 
\textbf{transport plan}\footnote{A general definition can be made for any pair of probability measures.} between $\mathbf{r}$ and $\mathbf{c}$ if $P$ has $\rr$ and $\cc$ as its marginals. 
Denote the set of all \textit{transport plans} between $\mathbf{r}$ and $\mathbf{c}$ by $U(\mathbf{r}, \mathbf{c})$. 
Further, let a non-negative $C=(C_{ij})_{n\times m}$ be the cost matrix, where $C_{ij}$ measures the cost of transportation between $r_i$ and $c_j$.


\cite{cuturi2013sinkhorn} proposed \textit{Entropy regularized Optimal Transport (EOT)}. 
EOT seeks an optimal transport plan $P^{(\lambda)}$
that minimizes the entropy regularized cost of transporting $\mathbf{r}$ into $\mathbf{c}$. For a parameter $\lambda >0$, 
\begin{equation} \label{eq:EOT}
P^{(\lambda)} = \argmin_{P \in U(\mathbf{r}, \mathbf{c})} \hspace{0.1in}\{\langle C, P\rangle - \frac{1}{\lambda} H(P)\},    
\end{equation}

\noindent where $\langle C, P\rangle =  \sum_{i\in \dataSpace, j\in \conceptSpace} C_{ij}P_{ij} $ is the
Frobenius inner product between $C$ and $P$, and $H(P) := \sum _{i}^{n}\sum_{j}^{m} P_{ij} \log P_{ij}$ is the \textit{entropy} of $P$.
$P^{(\lambda)}$ is called a \textbf{Sinkhorn plan} with parameter $\lambda$. 

\noindent \textbf{Sinkhorn scaling.} 
Sinkhorn plans can be computed efficiently via Sinkhorn scaling with linear convergence \citep{knight2008sinkhorn}.
$(\rr, \cc)$-\textit{Sinkhorn scaling (SK)} \citep{Sinkhorn1967} of a matrix $M$ is simply the iterated alternation of row normalization of $M$ with respect to $\rr$ and 
column normalization of $M$ with respect to $\cc$ (See Example~\ref{eg: apd-sk} in Supplementary Text). 
When marginal distributions are uniform, we sometimes call it \textit{Sinkhorn iteration}. It is shown in \cite{cuturi2013sinkhorn} that, 

\begin{prop}\label{prop: sk_for_sp} 
Given a cost matrix $C$, a Sinkhorn plan $P^{(\lambda)}$ of transporting $\mathbf{r}$ into $\mathbf{c}$ 
can be obtained by applying $(\rr, \cc)$-Sinkhorn scaling on $ P^{[\lambda]}$, 
where matrix $P^{[\lambda]}$ is defined by $P^{[\lambda]}_{ij} = e^{-\lambda \cdot C_{ij}}$, thus:

\begin{equation}\label{eq:sk_distance} 
  P^{(\lambda)} = \text{SK}(P^{[\lambda]}) \text{\hspace{0.15in} and \hspace{0.15in}} 
  P^{[\lambda]} := e^{-\lambda \cdot C} = (e^{-\lambda \cdot C_{ij}})_{n\times m}.   
\end{equation}

\end{prop}

Much more is known about EOT and SK (see \citep{idel2016review} and our Supplemental Text Section~\ref{apd:SK_features}). 

\subsection{Cooperative communication as optimal transport}\label{sec:unify_OT}


\textit{Cooperative communication} formalizes a single problem comprised of interactions between two processes: action selection (teaching) and inference (learning) \citep{Shafto2014,jara2016naive,goodman2016pragmatic,fisac2017pragmatic}.
The teacher and learner have beliefs~about hypotheses, which are represented as probability distributions. The process of teaching is to select~data that move the learner's beliefs from some initial state, to a final desired state. 
The process of learning is then, given the data selected by the teacher, infer the beliefs of the teacher. 
The teacher's selection and learner's inference incur costs. 
The agents minimize the cost to achieve their goals. 
Communication is successful when the learner's belief, given the teacher's data, is moved to the target distribution.
The connection between EOT and cooperative communication is established by modeling each process, teaching and learning, as a classical EOT problem.

\textbf{Framework.} 
Let $\conceptSpace$ be a hypothesis space and $\dataSpace$ be a data space. Denote the common ground between agents:
the shared priors on $\conceptSpace$ and $\dataSpace$ by $P_0(\conceptSpace)$ and $P_0(\dataSpace)$,
the shared initial matrix over $\dataSpace$ and $\conceptSpace$ by~$\M$ of size $\dataSize \times \conceptSize$. In general, up to normalization, $M$ is simply a non-negative matrix which also specifies the consistency between data and hypotheses ~\footnote{Data, $d_i$, are consistent with a hypothesis, $h_j$, when $M_{ij}>0$.}

In cooperative communication, a teacher's goal is to minimize the cost of transforming the shared prior over hypotheses $P_0(\conceptSpace)$ into shared prior over data points $P_0(\dataSpace)$. 
We define the teacher's cost matrix $C^T = (C^T_{ij})_{\dataSize \times \conceptSize}$ as: 
%
\begin{equation}\label{eq:teacher_cost}
C^T_{ij} = - \log P_{L}(h_j|d_i) + S_T(d_i),
\end{equation}
\noindent where $P_{L}(h_j|d_i)$ is the learner’s likelihood of inferring hypothesis $h_j$ given data $d_i$, and $S_T(d_i)$ is determined by the teacher's prior on the data $d_i$ which can be interpreted as teacher's expense of selecting data $d_i$. Thus, taking cooperation into consideration,
data $d$ is good for a teacher who wishes to communicate $h$ if $d$ has a low selecting expense and the learner assigns a high probability to $h$ after updating with~$d$. Symmetrically, a learner's cost matrix $C^L = (C^L_{ij})_{\dataSize\times \conceptSize}$ is defined as $ C^L_{ij} = -\log P_{T}(d_i|h_j) + S_L(h_j)$, where $P_{T}(d_i|h_j)$ is the teachers’s likelihood of choosing data $d_i$ given hypothesis $h_j$ and $S_L(h_j)$ is determined by the learner's prior on the hypothesis $h_j$.

\noindent \textbf{Optimal Planning.} A \textit{teaching plan} is a joint distribution $T = (T_{ij})$ over $\dataSpace$ and $\conceptSpace$,
where each element $T_{ij} = P_T(d_i, h_j)$ represents the probability of the teacher selecting $d_i$ to convey $h_j$.
Similarly a \textit{learning plan} is a 
joint distribution $L =(L_{ij})$, where $L_{ij} = P_{L}(d_i, h_j)$ represents the probability of the learner inferring $h_j$ given $d_i$.
Column normalization of $T$ and row normalization of $L$ are called \textit{conditional communication plans}.



Under our framework, the \textit{optimal cooperative communication plans} that minimize agents' costs on transmitting 
between $\conceptSpace$ and $\dataSpace$ are precisely the \textit{Sinkhorn plans} as in Equation~\eqref{eq:EOT}.
Hence, as a direct application of Proposition~\ref{prop: sk_for_sp}, we have 

\begin{prop}\label{prop: sk_op}
Optimal cooperative communication plans, $T^{(\lambda)}$ and $L^{(\lambda)}$, 
that achieve Sinkhorn plans of EOT with given $\lambda$, can be obtained through Sinkhorn Scaling on matrices determined by the common ground between agents: priors $P_0(\conceptSpace)$, $P_0(\dataSpace)$ and shared consistency matrix $\M$.
\end{prop}

Construction of optimal plans $T^{(\lambda)}$ and $L^{(\lambda)}$ using Prop.~\ref{prop: sk_op} is illustrated as follows.
Assume zero expense of data selection and uniform priors on both $\dataSpace$ and $\conceptSpace$.
A natural estimation of the learner is a naive learner whose learning plan is fully based on the shared $\M$. 
In this case, the teacher may approximate the learner's likelihood matrix by $L_0$, the row normalization of $\M$.
Hence the teacher's cost matrix defined in Eq.(\ref{eq:teacher_cost}) has the form $C^T = -\log L_0$.
As in Eq.(\ref{eq:sk_distance}), the optimal teaching plan with regularizer $\lambda$, denoted by $T^{(\lambda)}$, can be obtained by 
applying Sinkhorn iterations on $T^{[\lambda]}$, i.e.

\begin{equation}\label{eq:ot_teaching}
T^{(\lambda)}= SK(T^{[\lambda]}) = SK(e^{-\lambda \cdot C^T}) = SK(e^{\lambda \cdot \log L_0}) = SK(L_0^{[\lambda]}),
\end{equation}

\noindent where $L_0^{[\lambda]}$ represents the matrix obtained 
from $L_0$ by raising each element to the power of $\lambda$. Symmetrically, the optimal learning plan with regularizer $\lambda$, denoted by $L^{(\lambda)}$, can be reached by Sinkhorn iteration on $L^{[\lambda]} = e^{-\lambda \cdot C^L} = T_0^{[\lambda]}$, where $T_0$ is the column normalization of $\M$. 
Parameter $\lambda$ controls the agents' greediness towards deterministic plans, which is investigated in Section~\ref{sec: lambda_dynamic}.




\subsection{Unifying existing theories of cooperative communication} \label{sec:existing_models}
A wide range of existing cooperative models in pragmatic reasoning, social cognitive development and robotics can be unified as approximate inference for EOT. The major variations among these models are: depth of Sinkhorn scaling and choice of parameter $\lambda$. See a brief summary in Table~\ref{tab:unifying}

\textbf{Fully recursive Bayesian reasoning.} The first class is based on the classic Theory of Mind recursion, including \textit{pedagogical reasoning} \citep{Shafto2008,shafto2012learning,Shafto2014} and \textit{cooperative inference} \citep{YangYGWVS18,wang2018generalizing}. 
These models use fully Bayesian inference to compute the exact Sinkhorn plans (i.e. Sinkhorn scaling until convergence) 
for the case of $\lambda =1$. 
In more detail, these models emphasize that agents' optimal conditional communication plans, $T^{\star} =P_{T}(\dataSpace|\conceptSpace)$ and $L^{\star}= P_{L}(\conceptSpace|\dataSpace)$ should satisfy the following system of interrelated equations, each of which is in form of the Bayes's rule:

\begin{equation}\label{eq:CI}
 \LL = \frac{\TT \LLpri}{\LLmar} \hspace{0.2in} \TT = \frac{\LL \TTpri}{\TTmar}
\end{equation}

\noindent where $\LLmar$ and $\TTmar$ are the normalizing constants. The main theorem in \cite{YangYGWVS18} shows that assuming 
$\LLpri$ and $\TTpri$ are uniform priors over $\conceptSpace$ and $\dataSpace$, 
Eq.\eqref{eq:CI} can be solved using SK iteration on the shared matrix $M$. Hence coincide with Sinkhorn plans of EOT. Moreover, benefiting directly from the EOT framework, Prop.~\ref{prop: sk_op} implies and extends this result to \textit{arbitrary priors}:

\begin{prop}\label{prop:ci_is_ot} \footnote{All proofs are included in  Section~\ref{apd:eg_proof} of Supplementary Text (\textbf{ST}).}
Optimal conditional communication plans, $T^{\star}$ and $L^{\star}$, 
of a cooperative inference problem with arbitrary priors, can be obtained through Sinkhorn scaling. 
In particular, as a direct consequence, cooperative inference is a special case of the unifying EOT framework with $\lambda=1$.
\end{prop}



\textbf{One-step approximate inference.} The second class is based on human behaviors such as \textit{Naive Utility Calculus} \citep{jara2016naive,jern2017people}, \textit{Rational Speech Act} (RSA) theory \citep{goodman2016pragmatic,franke2016probabilistic} and \textit{Bayesian Teaching} \citep{Eaves2016b,Eaves2016c}, and recent advances in robotics and machine learning, such as \textit{machine teaching} \citep{zhu2013machine, zhu2015machine}, \textit{pedagogical interaction} \citep{ho2016showing,ho2018effectively} and \textit{value alignment} \citep{hadfield2016cooperative,fisac2017pragmatic,jara2019theory}. 
These models compute one or two steps of the Sinkhorn scaling, 
then approximate the Sinkhorn plans of EOT either with the resulting probability distribution or form a deterministic plan using argmax (See detailed demonstrations in Supplementary Text Sec.~\ref{apd:existing_models}). Greediness parameter $\lambda$ is fitted as hyperparameter for different applications. 
The EOT framework suggests in many cases, such approximations are far from optimal (illustrated in Fig.~\ref{fig:robust}) and are much more sensitive to agents' estimation of the other agent (see Sec.~\ref{sec: sk_robust}).

\begin{table}
 \caption{Unifying existing cooperative models by EOT framework}
    \label{tab:unifying}

    \centering
\begin{tabular}{ llll  }
 \toprule
 Example of Existing Models & Depth of SK & choice of $\lambda$ & Stochasticity\\
 \midrule

 Pedagogical Reasoning \citep{Shafto2014} & until converge & fit per data & probabilistic\\

 Cooperative Inference \citep{YangYGWVS18} & until converge   & 1 &  probabilistic \\

Bayesian Teaching \citep{Eaves2016c} & 1 step & 1 & probabilistic\\
Machine Teaching \citep{zhu2013machine} & 1 step & N.A. (argmax) & deterministic\\

Naive Utility Calculus \citep{jara2016naive} & 1 step & 1 & probabilistic \\

 RSA \citep{goodman2016pragmatic} &  1-2 steps  &  fit per data & probabilistic \\

Value Alignment \citep{fisac2017pragmatic} & 1 step & fit per data &  deterministic \\
\bottomrule
\end{tabular}
    \vspace{1pc}
   
\end{table}

\noindent \subsection{Connections to Information theory} 

Cooperative communication, like standard information theory, involves communication over a channel. It is therefore
interesting and important to ask whether there is a formal connection. 
The EOT formulation shows that the cooperative communication is closely related to lossy data compression in \textit{rate-distortion theory} as follows.

Let $X = \{x_i\}_{i=1}^{m}$ be the source (input) space, $Y = \{y_j\}_{j=1}^n$ be the receiver (output) space, 
$P_0(X)$ be a fixed prior on $X$ and $Q = P(y_j|x_i)$ be a compression scheme.
Denote the distortion between $x_i$ and $y_j$ by $d(x_i, y_j)$, which measures the cost of representing $x_i$ in terms of $y_j$.
The \textbf{distortion} of a given compression scheme $Q$ is defined to be: $D_Q(X, Y) = \sum_{i,j} P_0(x_i) \cdot P(y_j | x_i) \cdot d(x_i, y_j) =  \sum_{i,j}  P(x_i, y_i) \cdot d(x_i, y_j)$. The amount of information (bits per symbol) communicated through scheme $Q$ is measured by the \textbf{mutual information}, $\displaystyle I(X, Y) = H(X)+H(Y) - H(X, Y)$, where $H(X)$, $H(Y)$ and $H(X,Y)$ are entropy of $P_0(X)$, $P_0(Y)$ and $P(X,Y)$ respectively.
The classical \textit{Distortion-rate function}, formulates the problem of minimizing distortion while passing at most $R$-bit per input symbol of information, thus find:
\begin{equation} \label{eq:rate-distortion}
 Q^* =  \arginf_{Q} D_Q(X, Y)  \text{ subject to } I(X,Y) < R.
\end{equation}
EOT minimizes the communication distortion by replacing the hard constraint on mutual information in Eq.~\eqref{eq:rate-distortion} by a soft regularizer. 
Consider the case where $X = \conceptSpace$, $Y = \dataSpace$, 
EOT is the problem that among all the compression scheme (communication plans) satisfying $P_0(\conceptSpace) = \mathbf{c}$ and $P_0(\dataSpace) = \mathbf{r}$, 
find the optimal plan that minimizes the distortion subject to penalties on bits per symbol. 
The penalty level is controlled by $\lambda$.
Thus, in the notation of rate-distortion theory, Eq.~\eqref{eq:EOT} of EOT is equivalent to:
$\displaystyle P^{(\lambda)} = \arginf_{P \in U(\mathbf{r}, \mathbf{c})} D_P(\conceptSpace, \dataSpace) + \frac{1}{\lambda} I(\conceptSpace, \dataSpace)$.

\section{Analyzing models of cooperative communication} \label{sec:analysis}

\subsection{EOT is statistically and information theoretically optimal}\label{sec:OT_statsbest}

Optimal cooperative plans of EOT solves \textit{entropy minimization} with marginal constraints through Sinkhorn scaling.
Let $M$ be a joint distribution matrix over $\dataSpace$ and $\conceptSpace$.
Denote the set of all possible joint distribution with marginals $\rr = P_0(\dataSpace)$ and $\cc = P_0(\conceptSpace)$ by $U(\rr,\cc)$.
Consider the question of finding the approximation matrix $P^*$ of $M$ in $U(\rr,\cc)$ that minimizes 
its relative entropy with $M$: 

\begin{equation}\label{eq:KL1}
P^*= \arginf_{P \in U(\rr, \cc)} D_{\text{KL}}(P||M), \text{where } D_{\text{KL}}(P||M)  = \sum_{i,j} P_{ij} \ln \frac{P_{ij}}{M_{ij}},     
\end{equation}
The $(\rr,\cc)$-SK scaling of $M$~converges to $P^*$ if the limit exists \citep{csiszar1989geometric,franklin1989scaling}. 
We therefore directly interpret cooperative communication under EOT as minimum discrimination information for \textit{pairs} of interacting agents.

Sinkhorn scaling also arises naturally as a \textit{maximum likelihood estimation}. Let $\widehat{P}$ be the empirical distribution of i.i.d. samples from a true underlying distribution, 
which belongs to a model family.
Then the log likelihood of this sample set over a distribution $M$ in the model family is given by 
$n\cdot \sum_{ij} \widehat{P}_{ij} \log M_{ij}$, where $n$ is the sample size. 
Comparing with Eq.~\eqref{eq:KL1}, it is clear that
maximizing the log likelihood (so the likelihood) over a given family of $M$ 
is equivalent to minimizing $D_{\text{KL}}(\widehat{P}||M)$.
When 
the model is in the exponential family, 
the maximum likelihood estimation of $M$ can be obtained through SK scaling with empirical marginals \citep{darroch1972generalized,csiszar1989geometric}.
Therefore, EOT planning can also be viewed as the maximum likelihood belief transmission plan.

\subsection{Robustness to violations of common ground}\label{sec: sk_robust}

In EOT, for a fixed regularizer $\lambda$, optimal plans are obtained through SK scaling on a matrix determined by $M$ w.r.t. $\rr = P_{0}(\dataSpace)$ and $\cc = P_{0}(\conceptSpace)$. This can be viewed as a map $\Phi$, from $(M, \rr, \cc)$ to the SK limit,
where the \textit{Common ground} -- priors $P_{0}(\dataSpace)$ \& $P_{0}(\conceptSpace)$, 
and mappings from beliefs to data, $M$ -- represent the assumption that cooperating agents share knowledge of each others' beliefs.
However, it is implausible (even impossible) for any two agents to have exactly the common ground. 
We now investigate differentiability of EOT. 
This ensures robustness of the inference where agents' beliefs and mappings from beliefs to data differ, which shows the viability of cooperative communication in practice. 

Let $M^{\e_{1}}$, $\rr^{\e_{2}}$ and $\cc^{\e_{3}}$ be vectors obtained by varying elements of $M$, $\rr$ and $\cc$ at most by $\e_i$, where $\e_i>0$ quantifies the amount of perturbation.
We show that:
\begin{prop}\label{thm: sk_cont}
For any non-negative shared $M$ and positive marginals $\rr$ and $\cc$,
if $\Phi(M^{\e_1}, \rr^{\e_2}, \cc^{\e_3})$ and $ \Phi(M,\rr,\cc)$ exist, then
$\Phi(M^{\e_1}, \rr^{\e_2}, \cc^{\e_3})\to \Phi(M,\rr,\cc)$ as $M^{\e_1} \to M, \rr^{\e_2} \to \rr, \cc^{\e_3} \to \cc$. 
\end{prop}

\noindent Continuity of $\Phi$ implies that small perturbations on $M, \rr, \cc$, yield close solutions for optional plans. 
Thus cooperative communicative plans are robust to deviations from 
common ground between agents (see demonstrations in Sec.~\ref{sec:exp_perturb}). 
In particular, if agents empirically estimate relevant aspects of common ground, derived cooperative plans will stabilize as the sample size increases.

Moreover, deviations in common ground are repairable in EOT without recomputing communication plans. 
When restricted to positive distribution $M$, 
\cite{luise2018differential} shows that $\Phi (M, \rr,\cc)$ is in fact smooth on $\rr$ and $\cc$.
We further prove that $\Phi$ is also smooth on $M$.
Therefore, the following holds:
\begin{thm}\label{thm:smoothness} \footnote{General result on non-negative shared distributions is stated and proved in Supp.Text Section~\ref{apd: smooth}}
Let $\mathcal{M}$ be the set of
positive matrices of shape
$\dataSize \times \conceptSize$,
representing
all possible shared distributions, let $\Delta^{+}_{\dataSize}$ and $\Delta^{+}_{\conceptSize}$ 
be the set of all positive prior distributions over $\dataSpace$ and $\conceptSpace$, respectively.
Then $\Phi:\mathcal{M}\times\Delta_{\dataSize}^+\times\Delta_{\conceptSize}^+ \rightarrow \mathcal{M}$ is $C^{\infty}$.
\end{thm}

Theorem~\ref{thm:smoothness} guarantees that the optimal plans obtained through SK scaling 
are infinitely differentiable. 
Gradient descent can be carried out via \textit{Automatic Differentiation} as in \cite{genevay2017learning}.
We explicitly derive the gradient of $\Phi$ with respect to both marginals and $M$ analytically in Sec~\ref{apd:gradient} of Supp.Text.
Based on the derived closed form, 
we demonstrate that EOT agents can reconstruct a better cooperative plan using linear approximation once they realized the deviation from the previously assumed common ground 
in Sec.~\ref{sec:exp_linear_approx}.
In human communication, common ground is often inferred as part of the communication process \citep{luise2018differential,hawkins2018emerging}.
Thus, the differentiability and the gradient formula significantly increase
the flexibility and practicality of the EOT framework. 



\subsection{Instability under greedy data selection}\label{sec: lambda_dynamic}
We now explore the effect of $\lambda$ on EOT plans. To simplify notation, we focus on square matrices, similar analysis applies for rectangular matrices using machinery developed in \cite{wang2018generalizing}.

\begin{definition}\label{def:diag}
Let $A=(A_{ij})$ be an $n\times n $ square matrix and $S_n$ be the set of all permutations of $\{1, 2, \dots, n\}$.
Given $\sigma \in S_n$, the set $D^A_{\sigma}$ of $n$-elements $\{A_{1,\sigma(1)},\dots, A_{n,\sigma(n)} \}$ is called a \textbf{diagonal}
of $A$ determined by $\sigma$. If $A_{k\sigma(k)}>0$ for all $k$, we say that $D_\sigma^A$ is \textbf{positive}.
$D^A_{\sigma}$ is called a \textbf{leading diagonal} if the product $d^A_{\sigma}=\Pi_{i=1}^{n}A_{i, \sigma(i)}$, 
is the largest among all diagonals of $A$. 
\end{definition}

\begin{definition}\label{def:cross ratios}
Let $A,B$ be two $n\times n$ square matrices and $D^A_{\sigma}$ and $D^A_{\sigma'}$ be two diagonals of $A$ determined by permutations $\sigma, \sigma'$. Denote the products of elements on $D^A_{\sigma}, D^A_{\sigma'}$ by $d^A_{\sigma}, d^A_{\sigma'}$. 
Then $\CR(D^A_{\sigma}, D^A_{\sigma'})=d^A_{\sigma}/d^A_{\sigma'}$ is 
called the \textbf{cross-product ratio} between $D^A_{\sigma}$ and $D^A_{\sigma'}$. Further, let the diagonals in $B$ determined by the same $\sigma$ and $\sigma'$ be 
$D^B_{\sigma}$ and $D^B_{\sigma'}$. We say $A$ is \textbf{cross-ratio equivalent} to $B$, if $d_{\sigma}^A\neq 0 \Longleftrightarrow d_{\sigma}^B\neq 0$
and $\CR(D^A_{\sigma}, D^A_{\sigma'})=\CR(D^B_{\sigma}, D^B_{\sigma'})$ holds for 
any $\sigma, \sigma'$.
\end{definition}


Given $\M$, 
consider the EOT problem for the teacher 
(similarly, for the learner). 
Recall that, as in Eq.~\eqref{eq:ot_teaching}, the optimal teaching plan $T^{(\lambda)}$ is the limit of SK iteration of 
$L_0^{[\lambda]}$.
Note that the limits of SK scaling on $L_0^{[\lambda]}$ and $M^{[\lambda]}$ (obtained from $L_0$ or $M$ by raising each element to power of $\lambda$)
are the same as they are cross-ratio equivalent (shown in \cite{wang2018generalizing}). 
Therefore to study the dynamics of $\lambda$ regularized EOT solutions, we may focus on $M^{[\lambda]}$ and its Sinkhorn limit $M^{(\lambda)}$. 

\begin{figure}[t!]
    \centering
    \textbf{a.}\includegraphics[scale=0.35]{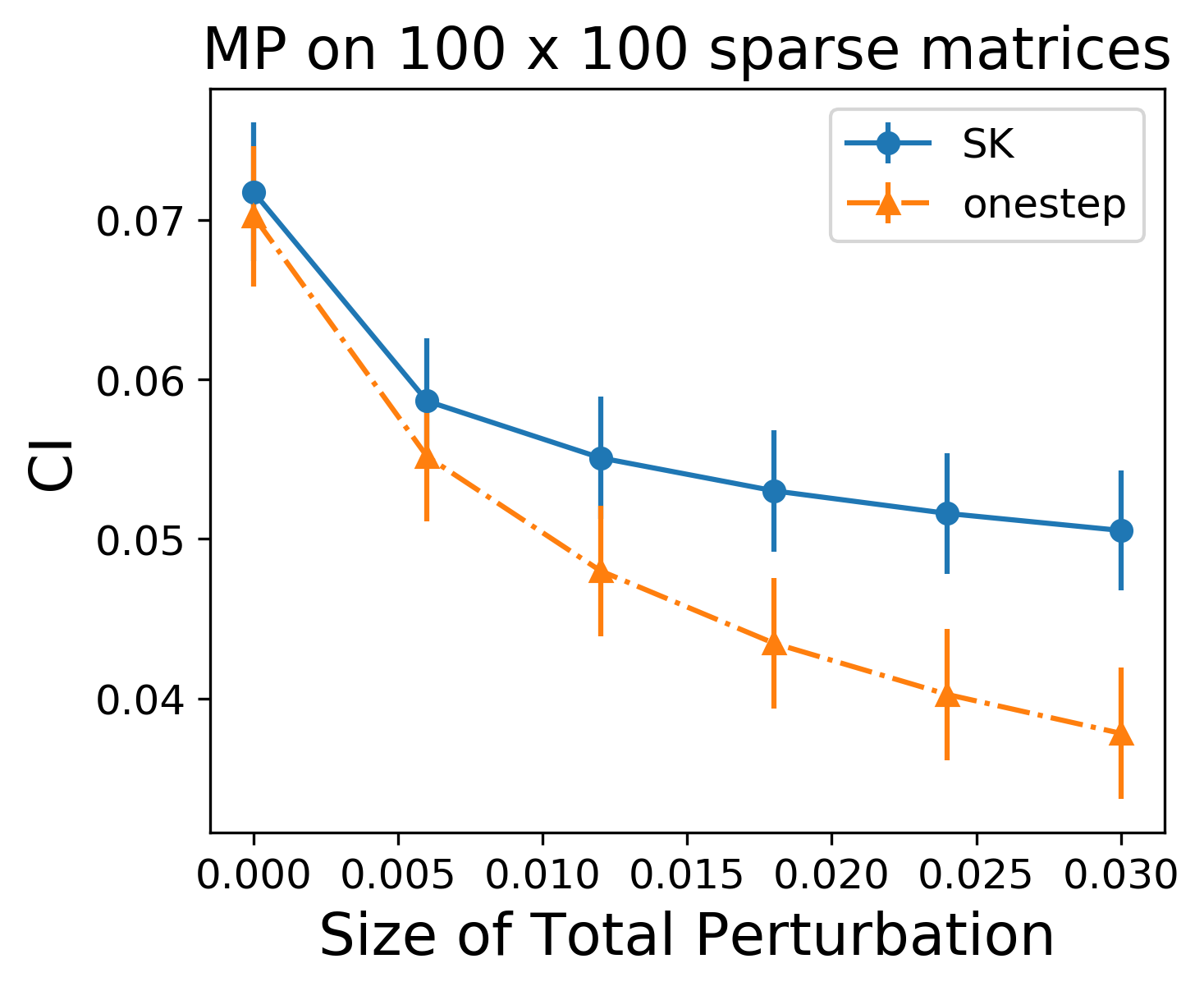}
    \textbf{b.}\includegraphics[scale=0.35]{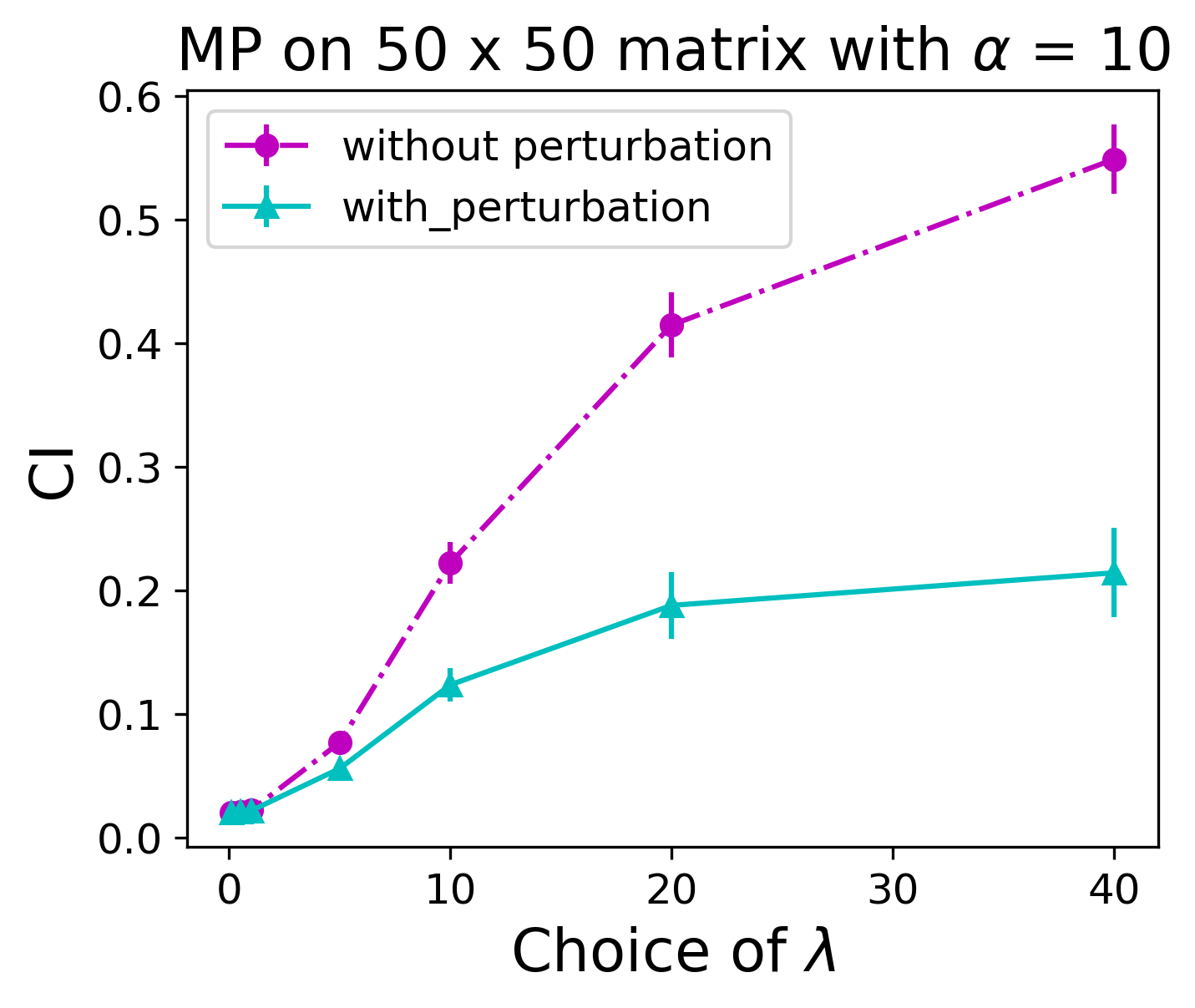}
    \textbf{c.}\includegraphics[scale=0.35]{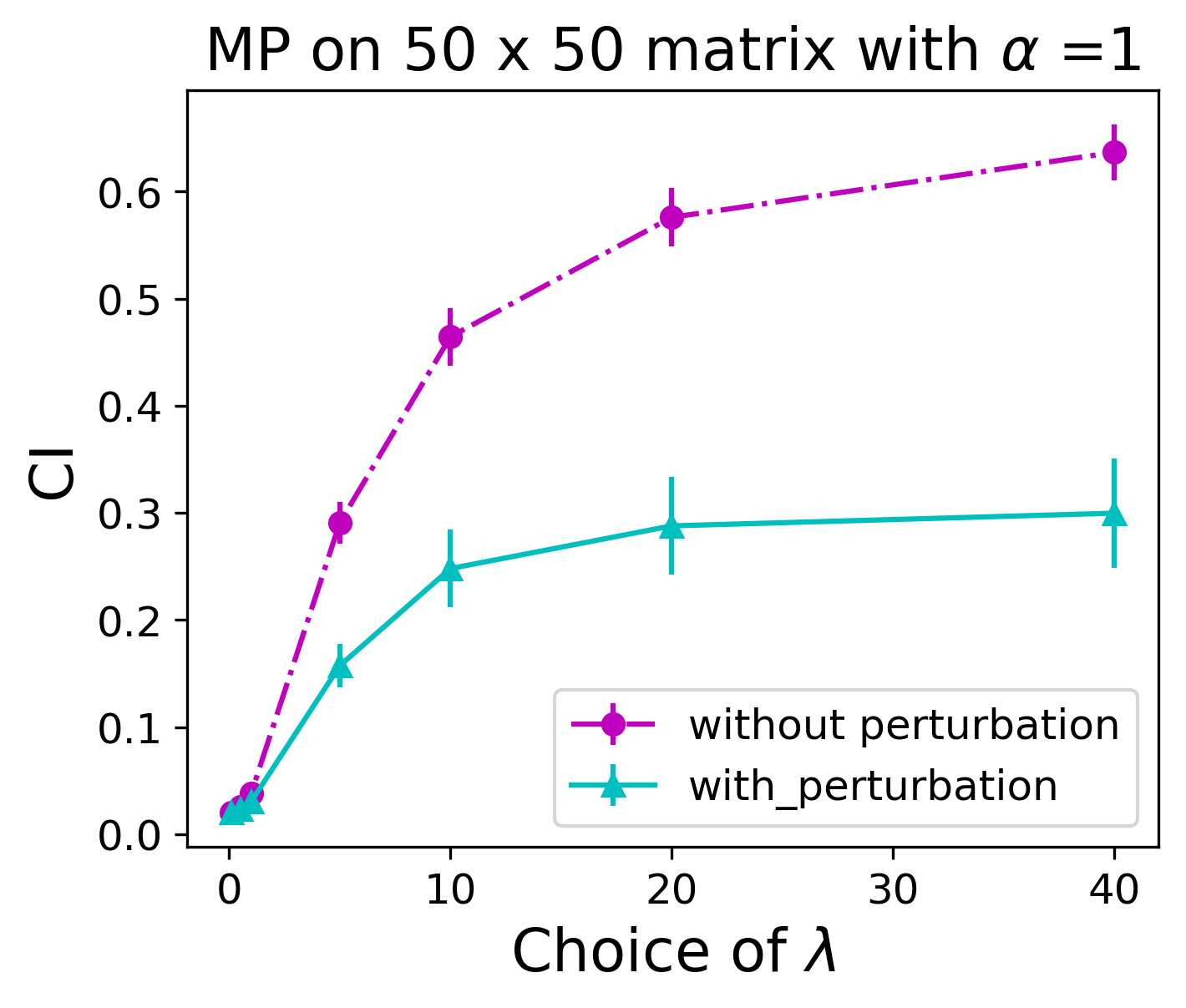}
    \textbf{d.}\includegraphics[scale=0.36]{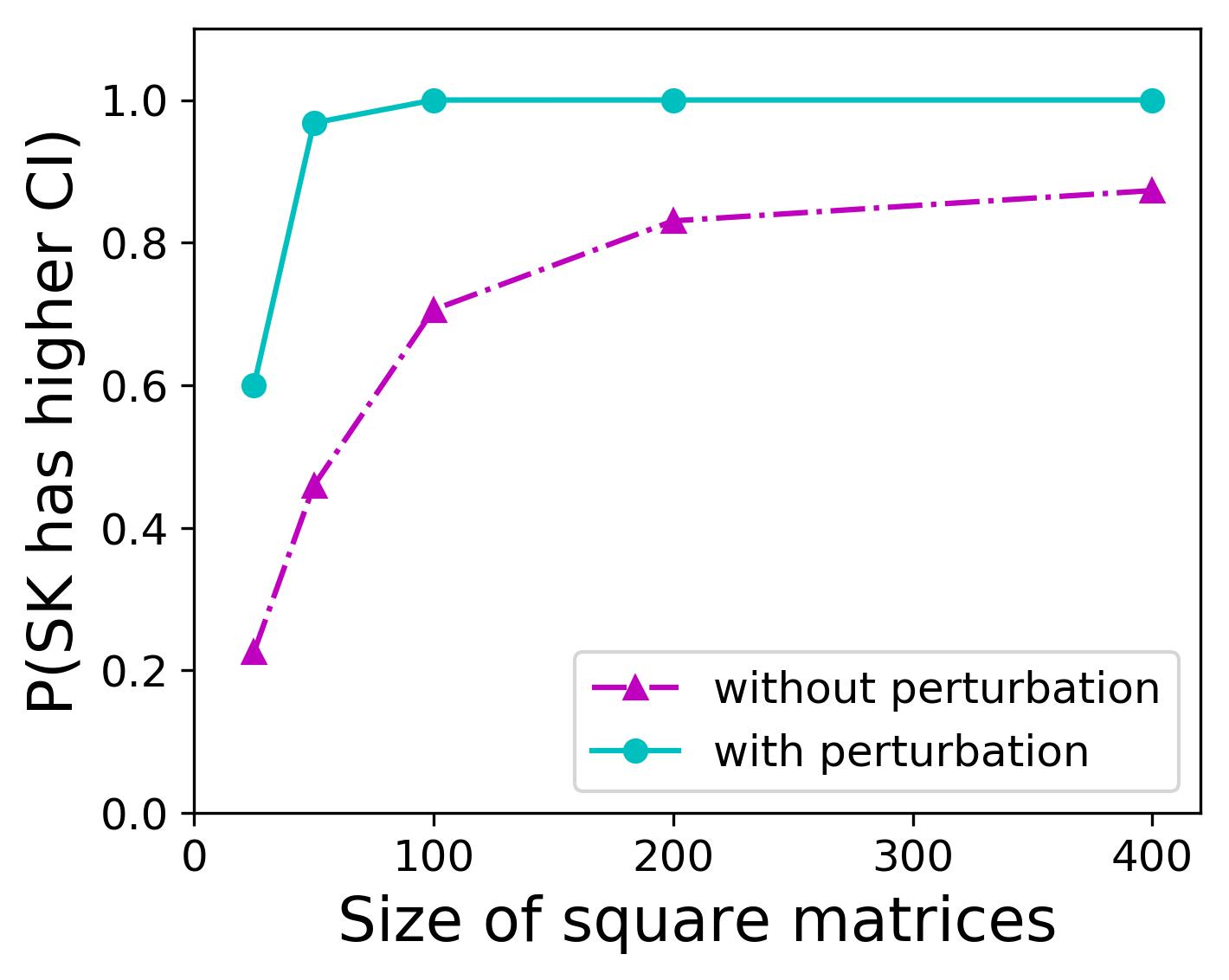}
    \textbf{e.}\includegraphics[scale=0.36]{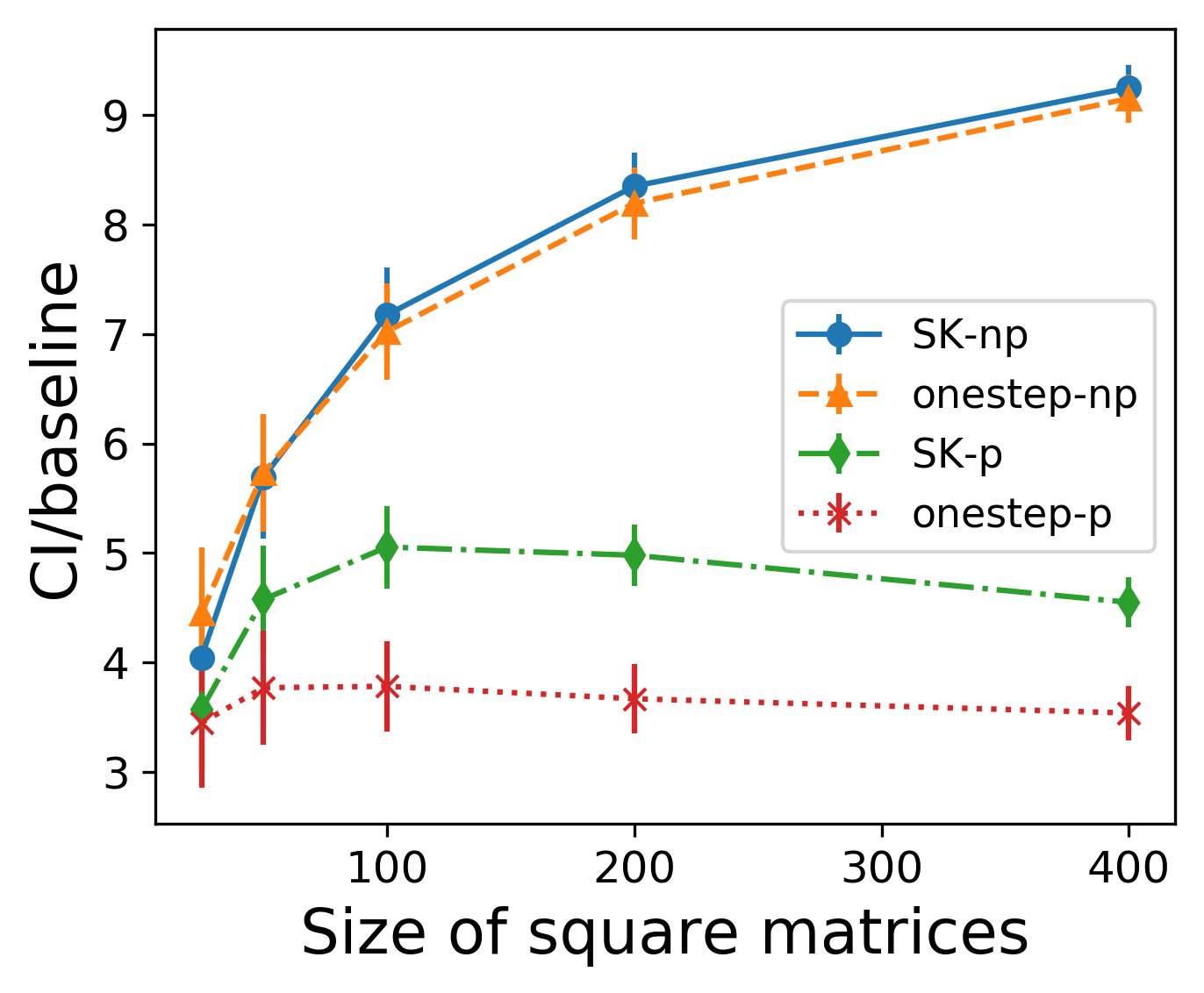}
    \textbf{f.}\includegraphics[scale=0.36]{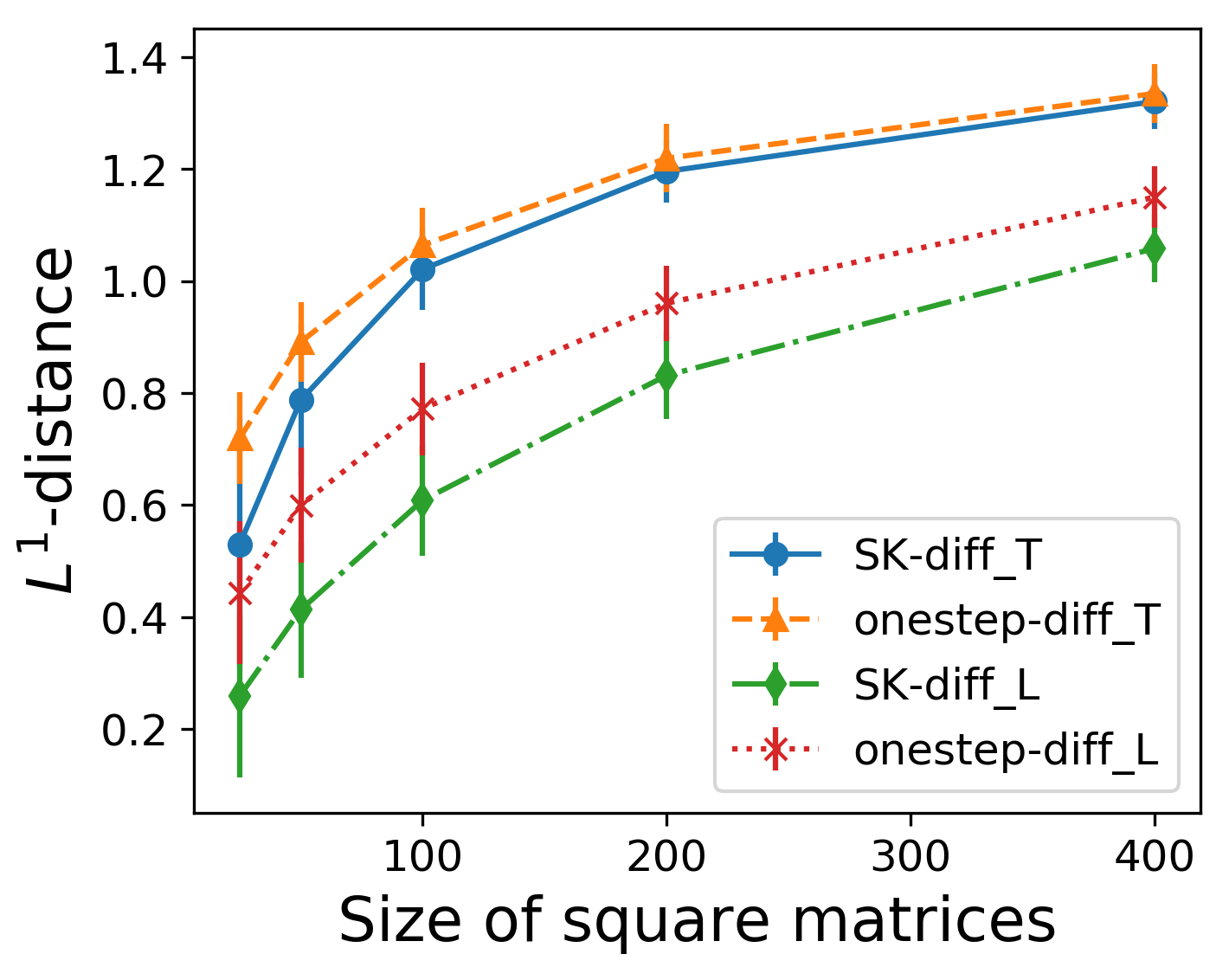}
    \caption{\textbf{a.} The Cooperative Index (CI) of Sinkhorn planning (SK) and its one step approximation (onestep) as total perturbation increases. 
    \textbf{b-c.} The average CI of Sinkhorn planning for $50\times 50$ matrices as $\lambda$ varies. \textbf{d-f.} $r = 0.03$, $\epsilon =1$, dimension of $M$ varies as shown in $x$-axis. \textbf{d.} The probability that CI of SK planning is higher than its one step approximation. \textbf{e.} The average communication effectiveness for SK and onestep with and without perturbations denoted by SK-p, onestep-p, SK-np, onestep-np accordingly. \textbf{f.} The average difference of the teacher's (and learner's) SK plans (and one-step approx.) before and after perturbations, measured by the $L^1$-distance.}
    \label{fig:robust}
\end{figure}

One extreme is when $\lambda$ gets closer to zero. 
If $\lambda \to 0$, $M^{[\lambda]}_{ij} = (M_{ij})^{\lambda} \to 1$ for any nonzero element of $M$.
Thus $M^{[\lambda]}$ converges to a matrix filled with ones on the nonzero entries of $M$,
and $M^{(\lambda)}$ converges to matrix $r^Tc$ if $M$ has no vanishing entries. 
Hence $M^{(\lambda)}$ reaches low communicative effectiveness as $\lambda$ goes to zero (demonstrated in Sec.~\ref{sec:exp_lambda} with Fig.~\ref{fig:robust}(b-c)).

The other extreme is when $\lambda$ gets closer to infinity.
In this case, assuming uniform priors, we show: 
\begin{prop}\label{prop: lambda}
$M^{(\lambda)}$ concentrates around the leading diagonals of $M$ as $\lambda \to \infty$.
\end{prop}
\noindent As $\lambda \to \infty$, the number of non-zero elements in $M^{(\lambda)}$ decreases. 
In the case when $M$ has only one leading diagonal, as $\lambda \to \infty$,
$M^{(\lambda)}$ converges to a diagonal matrix (up to permutation). Thus, it forms a bijection between $\dataSpace$ and $\conceptSpace$,
and achieves the highest effectiveness. 

The value of $\lambda$ causes variations on cross-ratios of $M^{(\lambda)}$, 
which affects the model's sensitivity to violations of common ground.
Since $M^{[\lambda]}$ and $M^{(\lambda)}$ are cross-ratio equivalent,
$M^{(\lambda)}$ has the same cross-ratio as the shared $M$ only when $\lambda =1$.
$M^{(\lambda\neq 1)}$ either exaggerates or suppresses the cross-product ratios of $M$, depending on whether $\lambda$ is greater or less than 1.
Hence, deviations on common ground are amplified by 
large $\lambda$, which reduces the communication effectiveness. 
Indeed, when deviation causes two agents have different leading diagonals in their estimations of $M$, 
their optimal plans will be completely mismatched as $\lambda \to \infty$ (See detail examples in Supp. Text Sec.~\ref{apd: add_epsilon}). 

\section{Experiments}\label{sec:exp}
We will now further illustrate properties of EOT through simulations.
Effectiveness of communication will be measured via the \textbf{Cooperative Index~(CI)} 
$\CI(T, L) := \frac{1}{\conceptSize}\sum_{ij}L_{ij}T_{ij}$ \citep{YangYGWVS18}. 
It ranges between $0$ and~$1$ and measures the communication effectiveness of a pair of plans $T$ and $L$.
Intuitively, $\CI(T, L)$ quantifies the effectiveness 
as the average probability that a hypothesis can be correctly inferred by a learner given the teacher's selection of data.



\subsection{Perturbation on common ground}\label{sec:exp_perturb}
In this section, we stimulate perturbations by Monte Carlo method to compare the robustness of the Sinkhorn planning and its one-step approximation.

\textbf{Basic Set-Up.} 
Assume a uniform prior on $\dataSpace$ and $\lambda =1$. Shared matrix $M$ and prior over $\conceptSpace$ are sampled from symmetric Dirichlet distribution with hyperparameter $\alpha =0.1$\footnote{The hyperparameter is set to be $0.1$ as sparse matrices are in general more sensitive to perturbations.}. Sample size is $10^6$ per plotted point. The scale of perturbations are controlled by two parameters: $r$, the percentage of elements to be perturbed; $\epsilon$, the magnitude of the perturbation on each element. For example, a $r=0.03$, $\epsilon = 0.5$ perturbation on $M$ represents that $3\%$ randomly selected elements of $M$ will be increased by $0.5*|M|_{\infty}$, where $|M|_{\infty}$ denotes the largest element of $M$. 
The communication effectiveness under perturbation is measured when one agent's common ground has varied.
Results on square matrices with perturbations on shared $M$ are presented here. Simulations on priors and rectangular matrices exhibit similar behaviors, see plots in Supp. Text Sec.~\ref{apd:sim}.

\textbf{Scaling Perturbation Size.}
We investigate effectiveness under increasing perturbation.
Matrices of size $100\times 100$ are sampled as described above. 
Fixing $r = 0.03$, $\epsilon$ is altered as in $[0,0.2,0.4,0.6,0.8,1]$. 
As shown in Fig.~\ref{fig:robust}a, effectiveness drops for the one-step approximation comparing to Sinkhorn plans 
when the magnitude of perturbation increases, illustrating robustness of EOT to violations of common ground.  

\textbf{Varying Matrix Dimension.} 
Fig.~\ref{fig:robust}d shows the effects of matrix dimension. 
We fix $r = 0.03$, $\epsilon =1$ and consider the dimension of $M$ in $[25, 50, 100, 200, 400]$. 
The probability that SK plans has higher CI than its one-step~approximation increases  
with the dimension of $M$. Moreover, the advantage of Sinkhorn planning is 
an effect that is increased in the presence of perturbations.

Fig.~\ref{fig:robust}e. plots the average communication effectiveness for SK Plans and its one-step approximation with and without perturbations. 
Since the communication problem naturally gets harder as the dimension of M increases, 
we use the ratio between $\CI$ and the dimensional baseline to measure the communication effectiveness, in stead of $\CI$.~\footnote{The dimensional baseline for a $N\times N$ matrix $M$ is set to be $1/N$, which is the probability that the learner infers the hypothesis teacher has in mind without communication.} 
Fig.~\ref{fig:robust}e. suggests that communication effectiveness is more stable for SK plans under perturbations.
Fig.~\ref{fig:robust}f. plots the average difference in $L^1$-distance of the teaching (and learning) plan before and after perturbations.
For instance, given $M$, denote the matrix after perturbation by $M_p$. Let $T^{sk}$, $T^{sk}_p$ be the teacher's SK plans obtained from EOT on $M$ and $M_p$ respectively. Their difference is measured as $|T^{sk}-T^{sk}_p|_{L_1}$. 
Fig.~\ref{fig:robust}f. shows that under perturbation, the deviations on SK plans are considerably smaller than its one-step approximations.

\subsection{Greedy selection of data}\label{sec:exp_lambda}
We investigate the effect of greedy parameter $\lambda$ on EOT when deviation occurs on agents' common ground. 
Fig.~\ref{fig:robust}b-c plot the average CI of Sinkhorn planning for $50\times 50$ matrices as $\lambda$ varies $[0.1,0.5,1,5,10,20,40]$. 
Fixing $r = 0.3$, $\epsilon =0.3$, the hyperparameter $\alpha$ of Dirichlet distribution for sampling $M$ is set to be 10 in Fig.~\ref{fig:robust}b, and 1 in Fig.~\ref{fig:robust}c. ($\alpha$ for $P_0(\conceptSpace)$ is set to be $10$ in both). 
The gap between the two curves expands in both Fig.~\ref{fig:robust}b-c, which illustrates that the robustness of EOT decreases as $\lambda$ grows.
As shown in Proposition~\ref{prop: lambda}, agents' optimal plan mainly concentrated on leading diagonals of their initial matrices. 
When deviation on $M$ causes mismatching leading diagonals for agents, $\lambda>1$ exaggerates the difference, hence the drop on the CI.
Notice that the rate of reduction of CI is more severe in \ref{fig:robust}b than \ref{fig:robust}c as $\lambda$ increases.
This is consistent with the model prediction (Section~\ref{sec: lambda_dynamic}) that under the same scale of perturbations, 
agents' plans are more likely to have variation on leading diagonals when element of the initial matrices are closer to evenly distributed.


\begin{figure}[h!]
    \centering
    \textbf{a.}\includegraphics[scale=0.5]{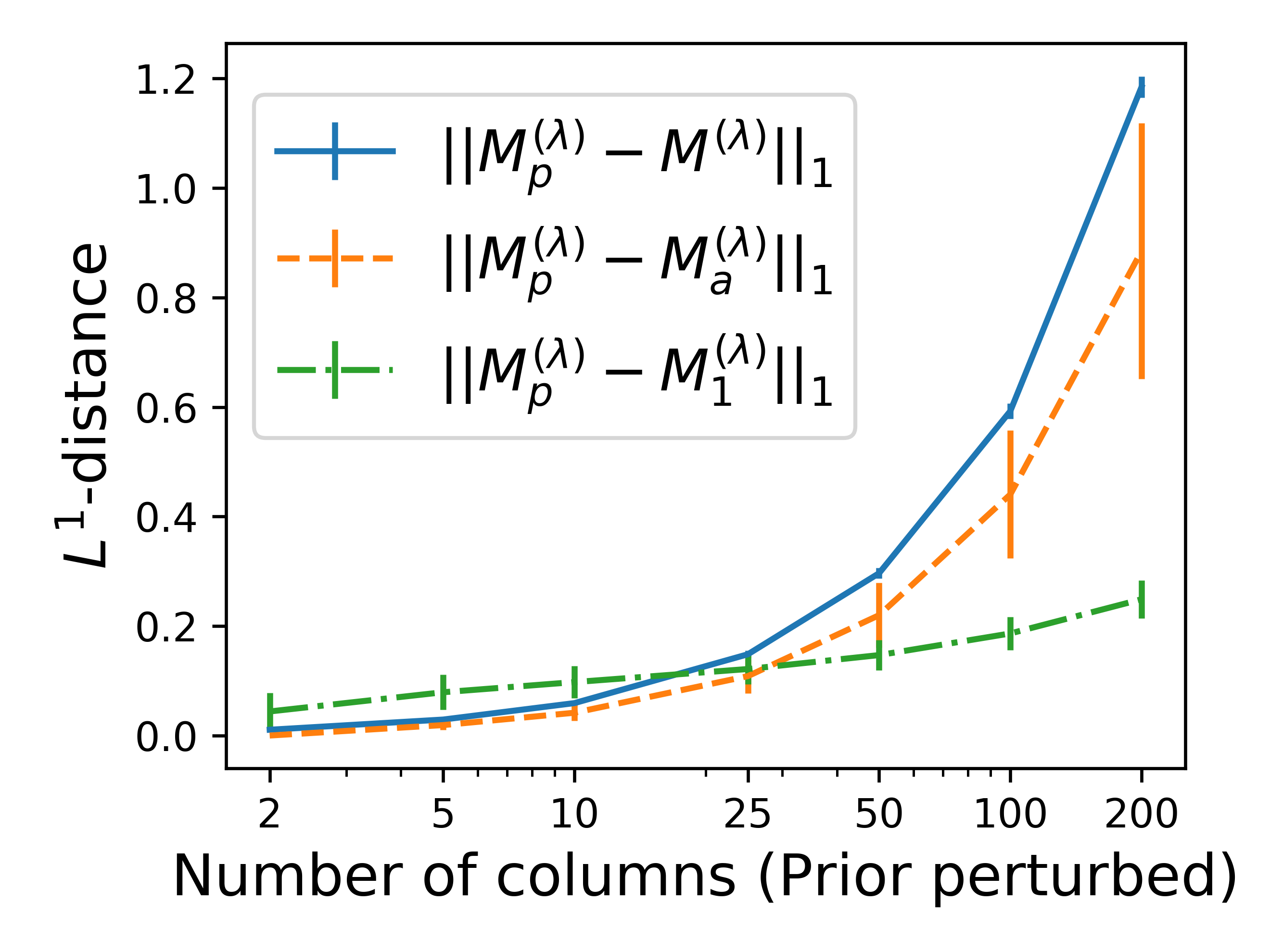}
    \hspace{0.2in}
    \textbf{b.}\includegraphics[scale=0.5]{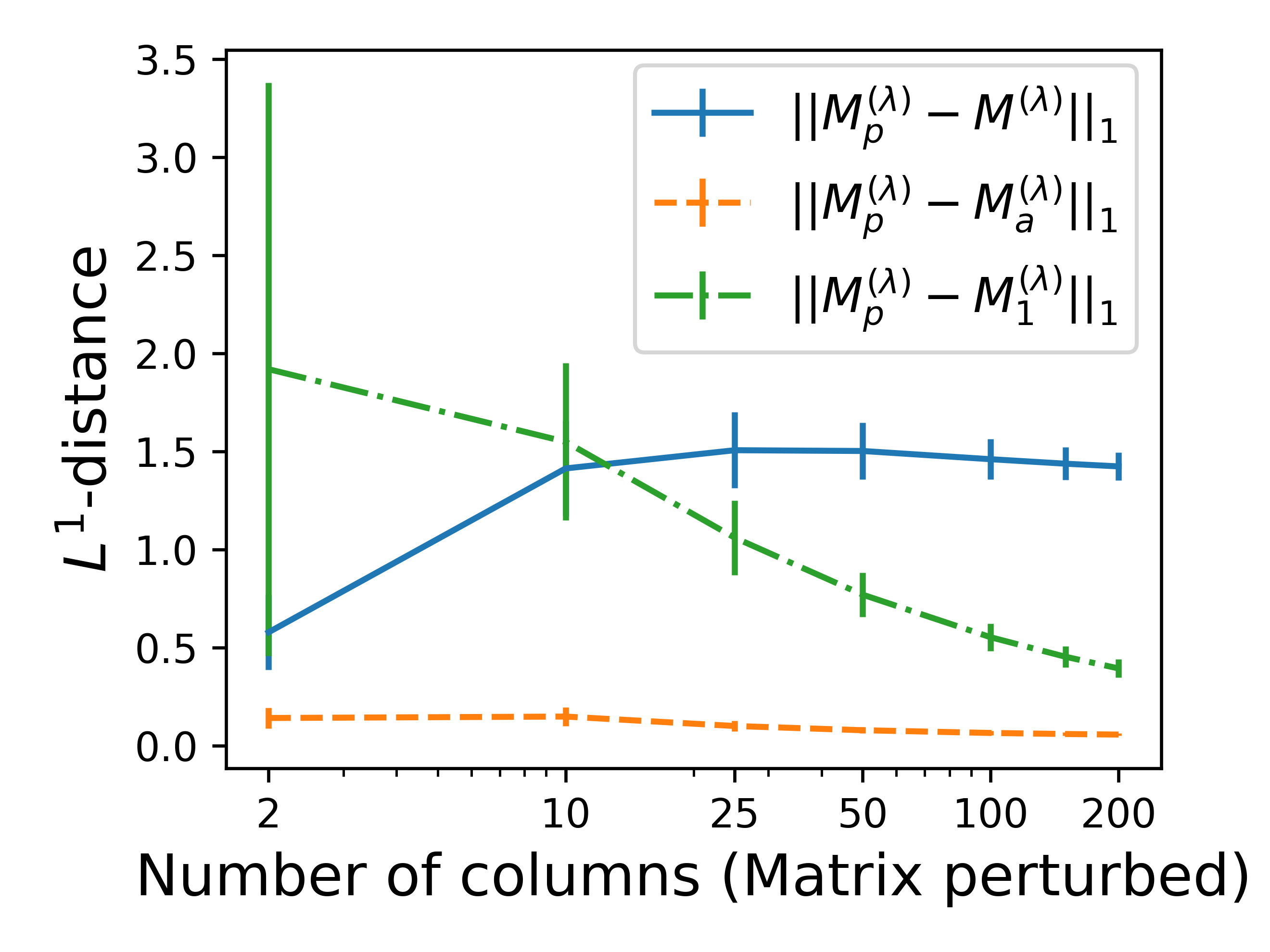}
    \caption{ Mean and stdev of the $L_1$-distance
    between SK plan $M_p^{(\lambda)}$ of $M_p$ and
    its three estimations: original SK plan $M^{(\lambda)}$ (blue),
    linear approximation $M_a^{(\lambda)}$ (orange), 
    and one step approximation $M_1^{(\lambda)}$ (green). }
    \label{fig:linear_approximation}
\end{figure}

\subsection{Linear approximation} \label{sec:exp_linear_approx}
The gradient guaranteed by Theorem ~\ref{thm:smoothness} allows 
online correction of deviations in common ground via linear approximation.  
Let $M_p$ be a deviation of $M$ obtained by perturbing elements of $M$.
To estimate the SK plan ($M^{(\lambda)}_p$) of $M_p$, 
we benchmark this linear approximation $M^{(\lambda)}_a=M^{(\lambda)}+\nabla_{\rr,\cc}\Phi\cdot\delta(\rr,\cc)+\nabla_{M}\Phi\cdot\delta M$ against the original SK plan $M^{(\lambda)}$, and the one-step approximation $M_1^{(\lambda)}$ of $M^{(\lambda)}_p$ \footnote{Thus, $M_1^{(\lambda)}$ is obtained from $M_p$ by one step Sinkhorn scaling.}. 
We use $L_1$-distance from each approximation to $M_p^{(\lambda)}$ to measure the error.

Fig.~\ref{fig:linear_approximation} shows the Monte-Carlo result
of $10^5$ samples. $\lambda=1$, $\rr$ and $\cc$ are uniform, and 
fix the number of 
rows to be $50$. Matrices, which differ in the number of columns
(labeled on x-axes, varying from $2$ to $200$), are
sampled so that each column follows Dirichlet distribution with parameter
$\alpha=1$. 
The perturbation on marginals are taken by adding $10\%$ to the 
sum of the first row while subtracting the same value from the sum
of the second row (Fig.~\ref{fig:linear_approximation}.a). 
The perturbation
on matrices is the same as in Sec.~\ref{sec:exp_perturb} with $r=0.03$
and $\epsilon=0.5$ (Fig.~\ref{fig:linear_approximation}.b). 
Linear approximation shows a modest effect for perturbations on the marginals, but is remarkably effective for perturbations on the matrix $M$.

\begin{figure}[t!]
    \centering
    \includegraphics[scale=0.4]{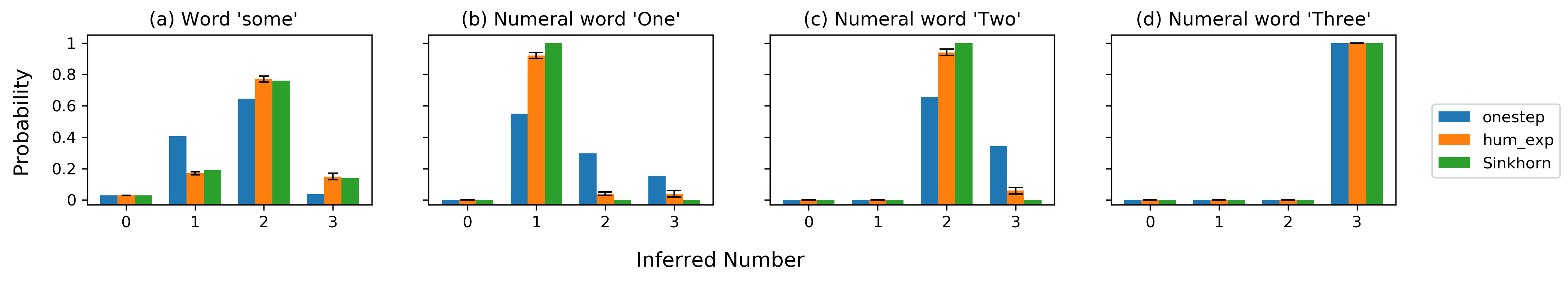}
    \caption{Green and blue bars plot the models' predictions regarding the learner's inference about the actual number of red apple based on the teacher's statement. The orange bars plot the empirical mean wager by the learner on each word state from \cite{goodman2013knowledge}.}
    \label{fig:number}
\end{figure}

\subsection{An application to human data} \label{sec:exp_apple}
We explore 
the following scenario from \cite{goodman2013knowledge}. Three apples, which could be red or green, are on a table.
The teacher looks at the table and make a statement quantifying the number of red apples such as "Some of the apples are red". 
The learner then infers the number of red apples based on the teacher's statement.
The hypothesis set $\conceptSpace = \{\text{`0',`1',`2',`3'}\}$ represents the true number of red apples,
and the data space $\dataSpace = \{\text{none, some, all}\}$ contains all the relevant quantifier words the teacher may choose.
Hence, the shared (unnormalized) consistency matrix for both agents is 
$\tiny{M = \begin{blockarray}{ccccc}
&\text{`0'} & \text{`1'} & \text{`2'}& \text{`3'}\\
\begin{block}{c (cccc)}
 \text{none}& 1 & 0 & 0 & 0\\
 \text{some}& 0 & 1 & 1 & 1 \\
 \text{all} & 0 & 0 & 0 & 1\\
\end{block}
\end{blockarray}}$. 
Both agents may estimate each other's likelihood matrix by normalizing $M$. 
The data were fit with a binomial prior distribution. Parameters for the one-step approximation as \citep{goodman2013knowledge} were base rate $0.62$, and $\lambda =3.4$ 
and for EOT were base rate $0.82$ (any choice of $\lambda$). Fig.~\ref{fig:number}(a) plots both models' predictions (i.e. learning plan) and the mean 
wager on the actual number of red apple by experimental participants, based on the teacher's statement \footnote{Human data are measured based on Fig.2 of \cite{goodman2013knowledge}.}. In this case, both models successfully capture that `some' implies `not all'.



We further compare EOT and its one step approximation on interpretation of numerals. 
The setting is the same as above, except after looking at the table, the teacher makes a numeric statement such as "Two of the apples are red".
Fig.~\ref{fig:number}(b-d) shows simulation results with priors over $\conceptSpace$ and $\dataSpace$ be uniform and $\lambda =1$.
Notice that the EOT plan is in fact the identity matrix $I_4$. It is both more consistent with the human behavior experiments,
and achieves the highest possible communicate effectiveness as $\CI(I_4, I_4) = 1$,
whereas the one-step approximation only has $\CI=0.5$.

\section{Conclusions}

Formalizing cooperative communication as Entropy regularized Optimal Transport, we show that cooperative communication is provably effective in terms of maximizing likelihood of belief transmission and is robust and adaptable to violations of common ground, with probabilistic reasoning optimizing the trade-off between effective belief transmission and robustness to deviations in common ground. 
Thus, claims regarding cooperative communication of beliefs between quite different agents, such as parents and children, speakers and listeners, teachers and learners, across cultures, or even between humans and machines, are mathematically well-founded. 
Our approach, based on unifying probabilistic and information theoretic models under Entropy regularized Optimal Transport, may lead to new formal foundations for theories of human-human and human-machine cooperation.

\newpage

\section*{Broader Impact}

The theoretical approach introduced in this paper unifies models that have been proposed in the literatures on human language, education, and human-robot interaction---domains with significant societal implications. Our analysis highlights conditions under which they may be robust to violations of assumptions, and through mathematical analysis of previously algorithmic proposals, provides a means by which we may understand and improve the robustness of these models. This provides a mathematical framework within which we may understand their safe and responsible use in applications. More generally, the field of machine learning has not traditionally considered possibility that humans are a collaborative partner both in generating the datasets of interest and in using model's predictions. The theory advanced herein is explicitly models this collaboration toward the goal of more effective human-machine teaming. Thus, while the contributions of the current work are primarily theoretical, there are potential positive implications in areas of society interest. 



\begin{ack}

This project was supported by DARPA grant HR00112020039 the content of the information does not necessarily reflect the position or the policy of the Government, and no official endorsement should be inferred. 

This material is based on research sponsored by the Air Force Research Laboratory and DARPA under agreement number FA8750-17-2-0146 and the Army Research Office and DARPA under agreement HR00112020039. The U.S. Government is authorized to reproduce and distribute reprints for Governmental purposes notwithstanding any copyright notation thereon.

This work was also supported by DoD grant 72531RTREP, NSF SMA-1640816, NSF MRI 1828528 to PS. 

\end{ack}

\bibliographystyle{plainnat}
\bibliography{references}

\appendix
\newpage
\section*{Supplementary Text of A mathematical theory of cooperative communication}
\section{Properties of Optimal Transport and Sinkhorn scaling}\label{apd:SK_features}

\begin{example}\label{eg: apd-sk}
\textbf{An application of Sinkhorn Scaling and Proposition~\ref{prop: sk_for_sp}.}

Let $\rr =\cc=(\frac{3}{8},\frac{5}{8})$, and the cost matrix be 
$\tiny C = \begin{pmatrix} \log 1 & \frac13\log 2 \\
  \frac23\log 2 & \log 1 \end{pmatrix}$. 
For $\lambda=3$, we may obtain $P^{(3)}$ by applying SK scaling on
$\tiny P^{[3]} = \begin{pmatrix} e^{-3\log 1} & e^{-3\cdot \frac{1}{3}\log 2} \\ e^{-3\cdot \frac{2}{3}\log 2} & e^{-3\log 1} \end{pmatrix} 
= \begin{pmatrix} 1 & 1/2\\ 1/4 & 1 \end{pmatrix} $, which proceeds as follows: 
(a) row normalizing $P^{[3]} $ such that each row sum equals $1$, giving
$\tiny \begin{pmatrix} 2/3 & 1/3\\ 1/5 & 4/5 \end{pmatrix}$;
(b) multiplying the first row by $3/8$ and second row by $5/8$ giving
$\tiny L_0 =\begin{pmatrix} 1/4 & 1/8\\ 1/8 & 1/2 \end{pmatrix}$. 
Then similarly, column normalization of $L_0$ with respect to $\cc$ outputs $\tiny T_1 = \begin{pmatrix} 1/4 & 1/8\\ 1/8 & 1/2 \end{pmatrix}$. 
As $L_0 = T_1$, the SK scaling has converged with $P^{(3)} = T_1$.
In general, multiple iterations may be required to reach the limit.
\end{example}

We now summarize some of important features about OT and SK.

Numerous results on SK iteration have been proved. 
For instance, assuming uniform marginal distributions, SK iteration of a square $M$ converges if and only if $M$ has
at least one positive diagonal \citep{Sinkhorn1967} and the limit must be a doubly stochastic matrix, which can 
be written as a convex combination of permutation matrices \citep{dufosse2016notes}. 
SK iteration can be viewed as a continuous map \citep{sinkhorn1972continuous}.
For positive matrices, we illustrate, this map is in fact smooth, in particular differentiable.
This allows to show that the unifying OT framework is robust to various perturbations on the common grounds 
and to derive precise gradient formula to recover (linear approximate) optimal communication plans (Section~\ref{sec: sk_robust}).

After \cite{Sinkhorn1967},  the convergence results regarding Sinkhorn scaling has further
developed in various fields (see survey \citep{idel2016review}). 
SK converges at a speed that is several orders of magnitude faster than other transport solvers \citep{cuturi2013sinkhorn,allen2017much}. 
Sinkhorn plans have been extensively applied in machine learning algorithms, for example in 
barycenter estimation \citep{altschuler2017near}, supervised learning \citep{frogner2015learning}, 
domain adaptation \citep{courty2017optimal} and training GANs \citep{arjovsky2017wasserstein}.

There is a strong geometric intuition that underlies SK scaling via the cross-product ratio (Definition~\ref{def:cross ratios}). 
Matrices converge to the same limit under SK scaling if and only if they are cross-ratio equivalent \citep{wang2018generalizing}.
The space $\mathcal{K}(M)$ formed by all matrices with the same cross-product ratios as $M$ is a special manifold \cite{fienberg1968geometry}. 
SK scaling moves $M$ along a path in $\mathcal{K}$ to $M^*$ --- the unique intersection between $\mathcal{K}$ and the 
manifold determined by the linear marginal conditions \citep{fienberg1970iterative}.

Preservation of cross-product ratios over SK scaling implies that
Sinkhorn Plans of EOT are invariant under cost matrices constructed for agents with different depths of SK.
For instance in the illustration of Proposition~\ref{prop: sk_op} of the main text, instead of being naive, a learner could also be pragmatic who
would reason about his estimation of the teacher's reasoning and interpret data accordingly using Bayes' rule,
i.e. proportional to elements of $L_1$ which is row normalization of $T_0$.
Denote the teacher's cost matrices based on $L_0$ and $L_1$ by $C^T_0$ and $C^T_1$ respectively.
Because both $L_0$ and $L_1$ are derived from $M$ by applying Sinkhorn iteration, they are cross-ratio equivalent. So they have the same SK limit, i.e.
Sinkhorn plans with respect to both $C^T_0$ and $C^T_1$ are the same. Thus, even though the teacher's estimation of the learner was not accurate, 
the teacher's plan is still optimal.
Indeed, optimal teaching plans are equivalent for any learning matrix that is cross-ratio equivalent to the common ground~$M$.

Strengthened by the rich theory of OT, our framework can be used to solve much broader questions.
For example, general existence of OT planning between two arbitrary probability measures over any probability spaces 
are well-studied~\citep{villani2008optimal}.
This provides us machinery to study cooperative communications between agents even when $\conceptSpace$ and $\dataSpace$ are continuous spaces.
Further existence of optimal communicative plans are guaranteed as general existence of optimal couplings.
Moreover, OT plannings enjoy many other desirable features such as: the optimality passes to subsets, convexity of OT distance,
which enables broader perspectives on approximate inference and computation of optimal plans.

\section{Unifying existing theories of cooperative communication} \label{apd:existing_models}


Existing models of cooperative communication can be unified as approximate inference for EOT. 
In this section, we demonstrate this point by expressing representatives of three broad classes of models as EOT. 

\subsection{Full recursive reasoning is EOT.}
Cooperative models that build on the classic Theory of Mind recursion are methods utilizing fully Bayesian inference. 
For instance, \textit{cooperative inference} \citep{YangYGWVS18,wang2018generalizing} and \textit{pedagogical reasoning} \citep{Shafto2008,Shafto2014,shafto2012learning}. 
To simplify exposition, we will focus on the theory of cooperative inference and illustrate how Bayesian inference models fit into our unifying EOT framework.

The core of cooperative inference between two agents is that 
the teacher's selection of data depends on what the learner is likely to infer and vice versa.
Let $P_{L_{0}}(h)$ be the learner's prior of hypothesis $h\in \conceptSpace$, 
$P_{T_{0}}(d)$ be the teacher's prior of selecting data $d\in \dataSpace$, 
$P_{T}(d|h)$ be the teacher's posterior of selecting $d$ to convey $h$ 
and $P_{L}(h|d)$ be the learner's posterior for $h$ given $d$. 
\textbf{Cooperative inference} emphasizes that agents' optimal conditional communication plans, $T^{\star} =P_{T}(\dataSpace|\conceptSpace)$ and $L^{\star}= P_{L}(\conceptSpace|\dataSpace)$ should satisfy the following system of interrelated equations 
for any $d\in \dataSpace$ and $h\in \conceptSpace$, where $\LLmar$ and $\TTmar$ are the normalizing constants:
\begin{equation}\label{eq:LT}
 \LL = \frac{\TT \LLpri}{\LLmar} \hspace{0.2in} \TT = \frac{\LL \TTpri}{\TTmar}
\end{equation}


\noindent Results in \cite{YangYGWVS18} indicates that assuming uniform priors on $\dataSpace$ and $\conceptSpace$, 
Eq.\eqref{eq:LT} can be solved using Sinkhorn iteration on the joint distribution $M$.
More generally, we show:

\begin{repprop}{prop:ci_is_ot} 
Optimal conditional communication plans, $T^{\star}$ and $L^{\star}$, 
of a cooperative inference problem with arbitrary priors, can be obtained through Sinkhorn scaling. In particular, as a direct consequence, cooperative inference is a special case of the unifying EOT framework with $\lambda=1$.
\end{repprop}

\subsection{One-step approximate inference} \label{sec:RSA}

Models in social cognitive development and pragmatic reasoning, including \textit{Naive Utility Calculus} \citep{jara2016naive,jern2017people}, \textit{Rational Speech Act} (RSA) theory \citep{goodman2013knowledge,goodman2016pragmatic,franke2016probabilistic} and \textit{Bayesian Teaching} \citep{Eaves2016b,Eaves2016c} and their extensions 
\citep{jara2015beliefs,baker2017rational,jara2015children,liu2017ten,kiley2013mentalistic,jara2015not,bridgers2016children,gweon2018order,gweon2014children,jara2017children,cohn2018incremental,ong2015affective,ong2019computational} approximate cooperation as a single step of recursion. 

For instance, RSA models the communication between a speaker and a listener, formalizing cooperation that underpins pragmatic language. A pragmatic speaker selects an utterance optimally to inform a naive listener about a world state. Whereas a pragmatic listener interprets an utterance rationally and infers the state using one step Bayesian inference. This represents a communicative process where a speaker-listener pair can be viewed as a teacher-learner pair with world states-utterances being hypotheses-data points, respectively.

RSA distinguishes among three levels of inference: a \textit{naive listener}, a \textit{pragmatic speaker} and a \textit{pragmatic listener} \citep{goodman2013knowledge}. A~naive listener interprets an utterance according to its literal meaning. That is, given a shared matrix $\M$,
the naive listener's probability of selecting $h_i$ given $d_j$ is the $ij$-th element of $L_0$, which is obtained by row normalization of $\M$.

A pragmatic speaker selects an utterance to convey the state such that maximizes utility. In particular, they pick $d_i$ to convey $h_j$ by soft-max optimizing expected utility,
\begin{equation}
 P_{\text{T}}(d_i|h_j) \propto e^{\alpha\, U(d_i;h_j)}, 
\label{eq:utility}  
\end{equation}
where utility is given by $U(d_i;h_j) = \log L_0(h_j| d_i) - S(d_i)$,
which minimizes the surprisal of a naive listener when inferring $h_j$ given $d_i$ with an utterance cost $S(d_i)$.
This formulation is the same as one step of SK iteration in EOT framework (see Eq.\eqref{eq:sk_distance} and Eq.\eqref{eq:teacher_cost}) where $C^T= -U(d;h)$, $\lambda= -\alpha$. 


Next, a pragmatic listener reasons about the pragmatic speaker and infers the hypothesis using Bayes rule,
\begin{align}
P_{{L}}(h_j|d_i) \propto P_{{T}}(d_i|h_j) P_{L}(h_j), \label{eq:Bayes}   
\end{align}
Here $P_{\text{T}}(d_i|h_j)$ represents the listener’s reasoning on the speaker's data selection and $P_\text{L}(h_j)$ is the learner's prior. 
This is again one step recursion of EOT framework of $\lambda=1$.

As described above, teaching and learning plans in RSA are one-step approximations of the Sinkhorn plans. 
EOT framework suggests that in many cases, such approximations are far from optimal. For example, 
world states are often referred at many levels of specificity by human agents \citep{graf2016animal,hawkins2018emerging}, which yield a \textit{upper triangular} joint distribution matrix. EOT would output a \textit{diagonal matrix} as optimal plan which achieves the highest communication effectiveness,
whereas cooperative index of one step approximation is much lower. 
Furthermore, one-step approximation plans are much more sensitive to agents' estimation of the other agent.
For instance, a pragmatic speaker's teaching plan is tailored for a naive listener, in contrast the optimal plan 
obtained through fully recursion is stable for any listener derived from the same common ground. 

\subsection{Single-step argmax approximation}\label{sec:argmx}
Many recent advances in robotics involve artificial agents that implement human-like inverse planning \citep{fisac2017pragmatic,jara2019theory}, 
such as simple or structured desire inference \citep{baker2009action,velez2017interpreting,reddy2018you}, path and motion planning \cite{kim2016socially, dragan2013legibility}, pedagogical interaction \citep{ho2016showing,ho2018effectively} and value alignment \cite{hadfield2016cooperative,milli2017should}. 
In cooperative inverse reinforcement learning, instead of selecting acts probabilistically, the maximum probability action is selected.
For example, \citep{fisac2017pragmatic} 
introduces \textit{Pragmatic-Pedagogic Value Alignment}, a framework that is grounded in empirically validated cognitive models related to pedagogical teaching and pragmatic learning. 


Pragmatic-pedagogic value alignment formalizes the cooperation between a human and a robot who perform collaboratively with the goal of achieving the best possible outcome according to an objective. The true objective however is only known to the human. 
The human performs pedagogical actions to teach the true objective to the robot. After observing human's action, the robot, who is pragmatic, updates his beliefs and perform an action that maximizes expected utility. 
The human, observing this action, can then update their beliefs about the robot's current beliefs and choose a new pedagogic action. 
Denote actions by $d$ and objectives by $h$. We can see that when the human performs the action they act as a teacher 
and when robot is performing the action it is vice versa. 



In particular, the pedagogic human selects an action $d_i$ to teach the objective $h_j$ according to Eq.~\eqref{eq:utility},
where $U$ is the utility that captures human's best expected outcome.
As described in Section \ref{sec:RSA}, this is equivalent to a single step recursion in the EOT framework. 

Denote the robot's prior belief distribution on the objectives by $P_\text{R}(h_j)$.
The robot interprets the human's action $d_i$
rationally and updates his beliefs about the true objective using Bayes rule as Eq.~\eqref{eq:Bayes}.
Then acting as a teacher, the robot chooses an action that maximizes the human's expected utility using argmax function:

$ \hspace{1.5in}\displaystyle P_\text{R}(d_i)=\argmax_{d_R} \sum_{d_H, h_j}U(d_R, d_H;h)\cdot P_\text{R}(h_j)  $

\noindent where, $d_R$ denotes the robot's actions and $d_H$ denotes the human's actions. Unlike in human communication \citep{Eaves2016b,Eaves2016c} where the plans are chosen proportionally to a probability distribution, here the robot chooses a deterministic action using argmax function. 

As described above, inverse planning in robotics is modeled by computing a single step of Sinkhorn iteration and selecting the action that maximizes the outcome. 
Unlike full recursive reasoning is EOT, which tends to select the leading diagonal of the common ground $M$ as $\lambda \to \infty$ (Proposition~\ref{prop: lambda}), inverse planning methords like pragmatic-pedagogic value alignment selects the maximal element in each column of $M$, which is not even guaranteed to form a plan to distinguish every hypothesis. 
Hence a big concern of such argmax method is that for large hypothesis spaces, multiple hypotheses may reach argmax on the same data which lead to low communication efficiency. Further, continuity is generally lost for deterministic methods as argmax, which reduces the models' robustness comparing to EOT.


In summary, EOT framework unifies existing models of cooperative communication in social cognitive development, pragmatic reasoning and robotics with cooperative agents for specific missions and inference with different Sinkhorn iteration depths. This unification not only allows one to draw strong comparison of the relative merits and predictions of different theories, but also establish a potential toolbox for one to design assignment tailored models, which could achieve the best balance between efficiency and accuracy.

\section{Further discussion on Sensitivity for large $\lambda$}\label{apd: add_epsilon}

\begin{figure}[!ht]
    \centering
    \includegraphics[scale=0.4, trim=10 10 10 40, clip]{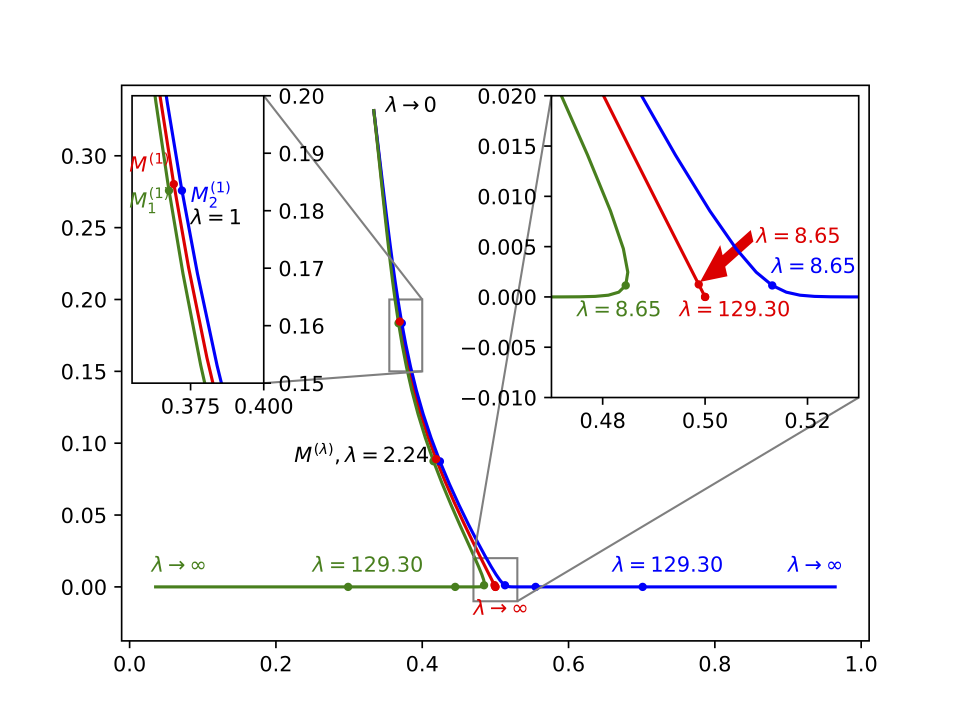}
    \caption{Lost of Continuity when $\lambda\rightarrow\infty$}
    \label{fig:lost_of_continuity}
\end{figure}

Sensitivity to perturbations is a concern as $\lambda \rightarrow \infty$. 
Figure~\ref{fig:lost_of_continuity} demonstrates an example where a slight variation on the initial matrices $M_1$ and $M_2$ can result a huge difference 
on $M_1^{(\lambda)}$ and $M_2^{(\lambda)}$ as $\lambda$ approaches infinity. 
The figure plots the Sinkhorn plans derived from $M_i^{(\lambda)}$ with the starting 
matrices $M_1$, $M_2$ differing from $M$ only by $2\%$ on their $l^\infty$-distance.
However, in this particular case, the 
change makes a huge difference: $M$ has two leading diagonals, while the perturbed $M_1$ and $M_2$ of $M$ enhanced one for each, making each $M_1$ and $M_2$ has only one leading diagonal. 
When $\lambda$ approaches zero, all products of diagonals tends to be the same, thus the curves (\textcolor{red}{red} for $M$, \textcolor[rgb]{0.06,0.512,0.006}{green} for $M_1$ and \textcolor{blue}{blue} for $M_2$)
converges to a common limit point, the uniform matrix.
But as $\lambda$ increases, the leading diagonals overwhelm other diagonals, and results in a fixed divergence on the limit when $\lambda\rightarrow\infty$.
Therefore, in this case, no matter how slight the changes are, as long as they modify the set of leading diagonals, there will be a fixed difference on the limits when $\lambda\rightarrow\infty$ according to the leading diagonals. Thus, $M^{(\infty)}$ is no longer continuous on the initial matrix $M$.

In particular, as $\lambda$ increases, the cooperative index, $\CI(M^{(\lambda)}_1,M^{(\lambda)}_2)$, between two agents with initial matrix $M_1$ and $M_2$ will be very small, even zero, if there is no overlapping positive element between $M^{(\lambda)}_1$ and $M^{(\lambda)}_2$ whereas $\CI(M^{(1)}_1,M^{(1)}_2)$ is bounded from below by the reciprocal of the number of diagonals of $M$.

\begin{example}
Assume that the teacher has the accurate 
$\tiny M = \begin{pmatrix} 1 & 5 &0 \\
   0& 1& 6 \\ 0&0&1\end{pmatrix}$. 
For any $\lambda$, the optimal teaching plan $T^{(\lambda)}=I_3$.
Suppose the learner gets constant noise of size $0.1$ in the position of $M_{31}$.
When $\lambda =1$, the learner's initial matrix  is
$\tiny L^{[\lambda=1]} = \begin{pmatrix} 1 & 5 &0 \\
   0& 1& 6 \\ 0.1 &0&1\end{pmatrix} $, 
the corresponding optimal plan is $\tiny  L^{(\lambda=1)} = \begin{pmatrix}  0.41& 0.51&0 \\
  0&  0.41&  0.51 \\ 0.51 &0&0.41\end{pmatrix} $ and
$\CI(T^{(\lambda)}, L^{(1)}) = 0.41$.
Similarly when $\lambda =2$, we have
$\tiny L^{[\lambda=2]} = \begin{pmatrix} 1 & 25 &0 \\
   0& 1& 36 \\ 0.01 &0&1\end{pmatrix} $, $\tiny  L^{(\lambda=2)} = \begin{pmatrix} 0.25 & 0.75&0 \\
  0&  0.25& 0.75 \\  0.75& 0& 0.25\end{pmatrix} $ and
$\CI(T^{(\lambda)}, L^{(2)}) = 0.25$.
Furthermore, as $\lambda \to \infty$, $\tiny L^{(\lambda) } \to \begin{pmatrix} 0 & 1&0 \\
  0&  0& 1 \\  1& 0& 0\end{pmatrix} $ and 
$\CI(T^{(\lambda)}, L^{(\lambda)}) \to 0$ . 
Thus, in this case communication efficiency is completely vanished due to deviations between the teacher and learner are exaggerated by greedy selection of examples.
\end{example}

\section{Simulations} \label{apd:sim}

\subsection{Perturbation on common ground and Greedy selection of data} \label{apd:sub:TI}

\begin{figure}[h!]
    \centering
    \textbf{a.}\includegraphics[scale=0.4]{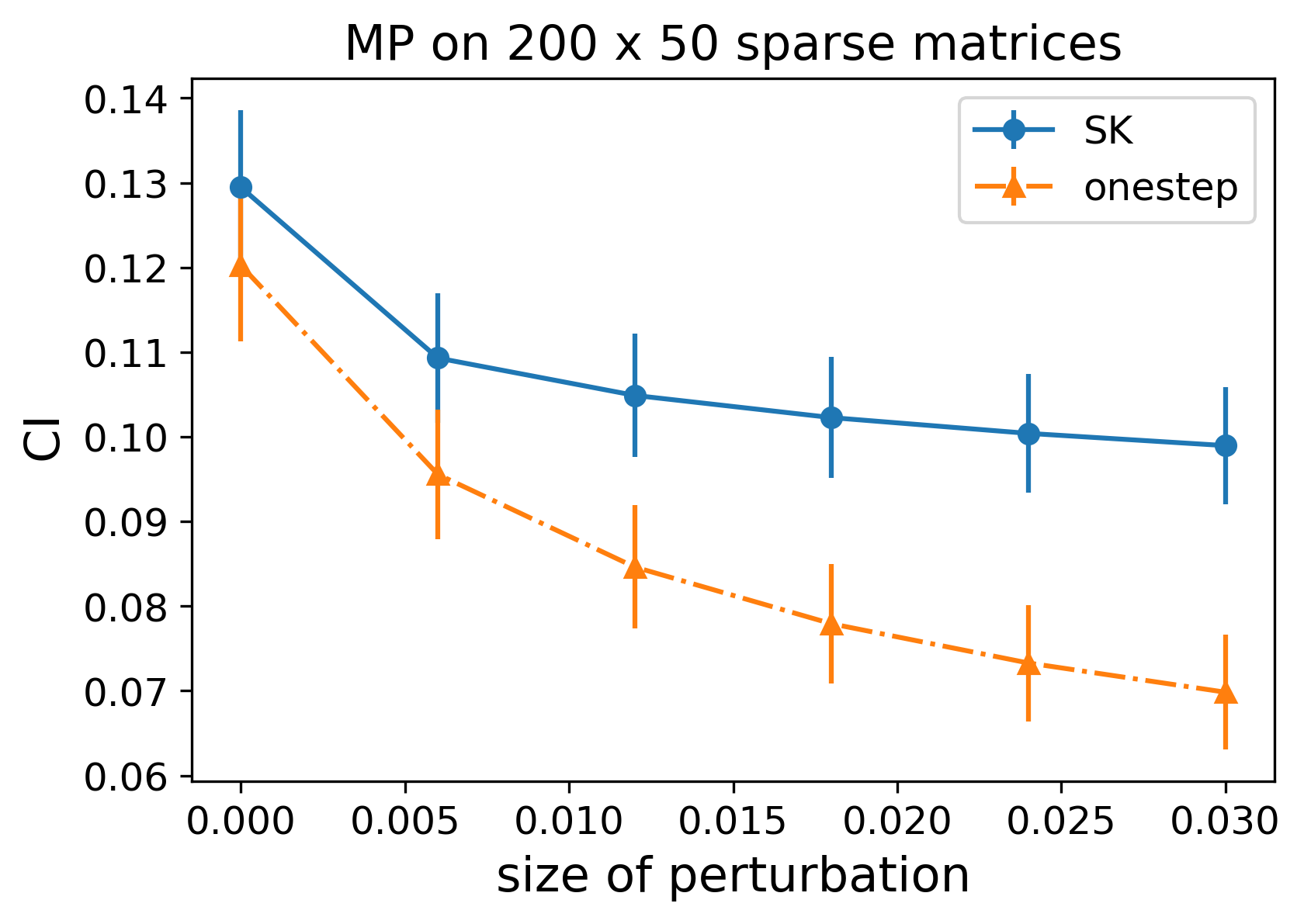}
    \textbf{b.}\includegraphics[scale=0.4]{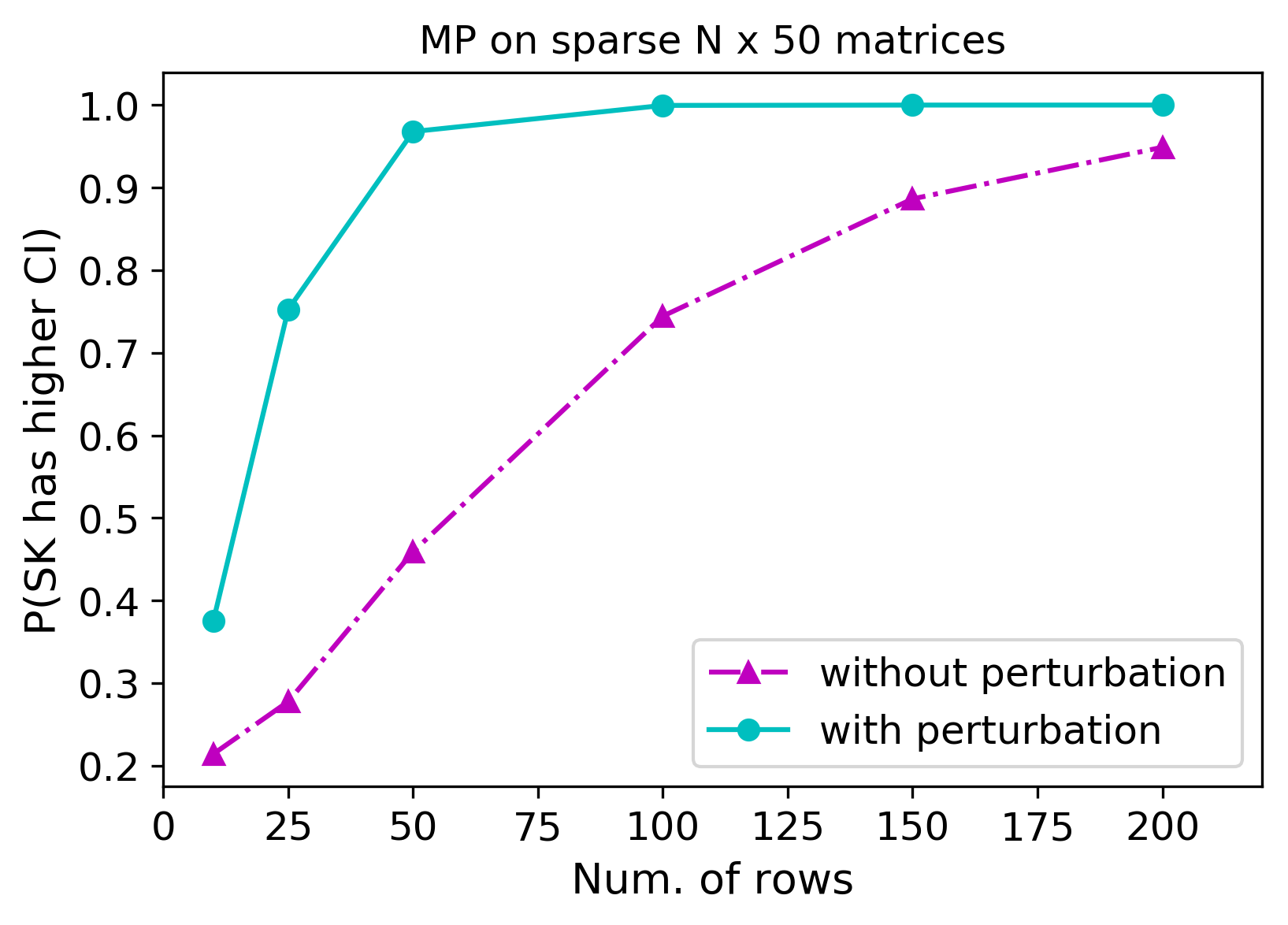}
    \textbf{c.}\includegraphics[scale=0.4]{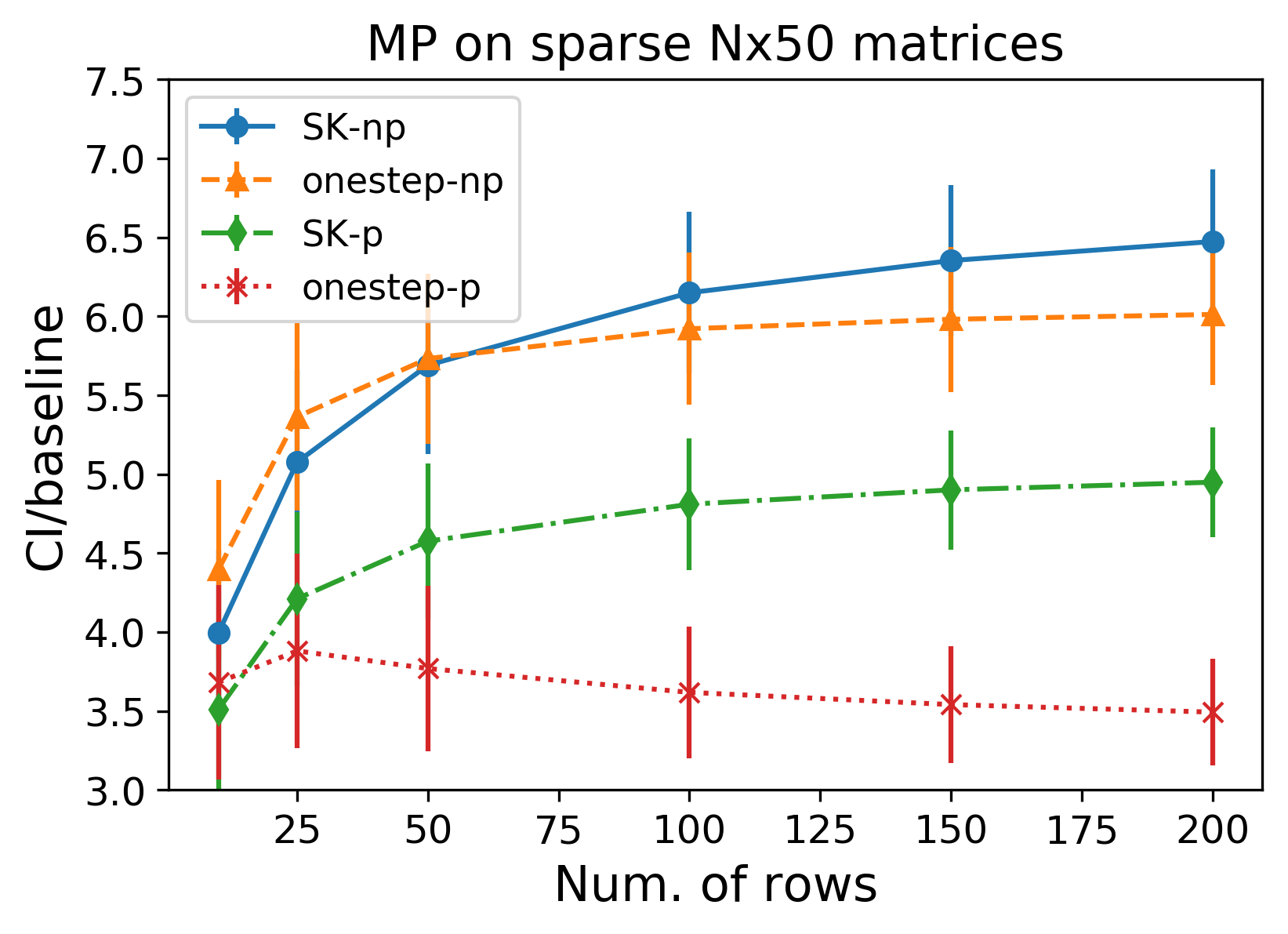}
    \textbf{d.}\includegraphics[scale=0.4]{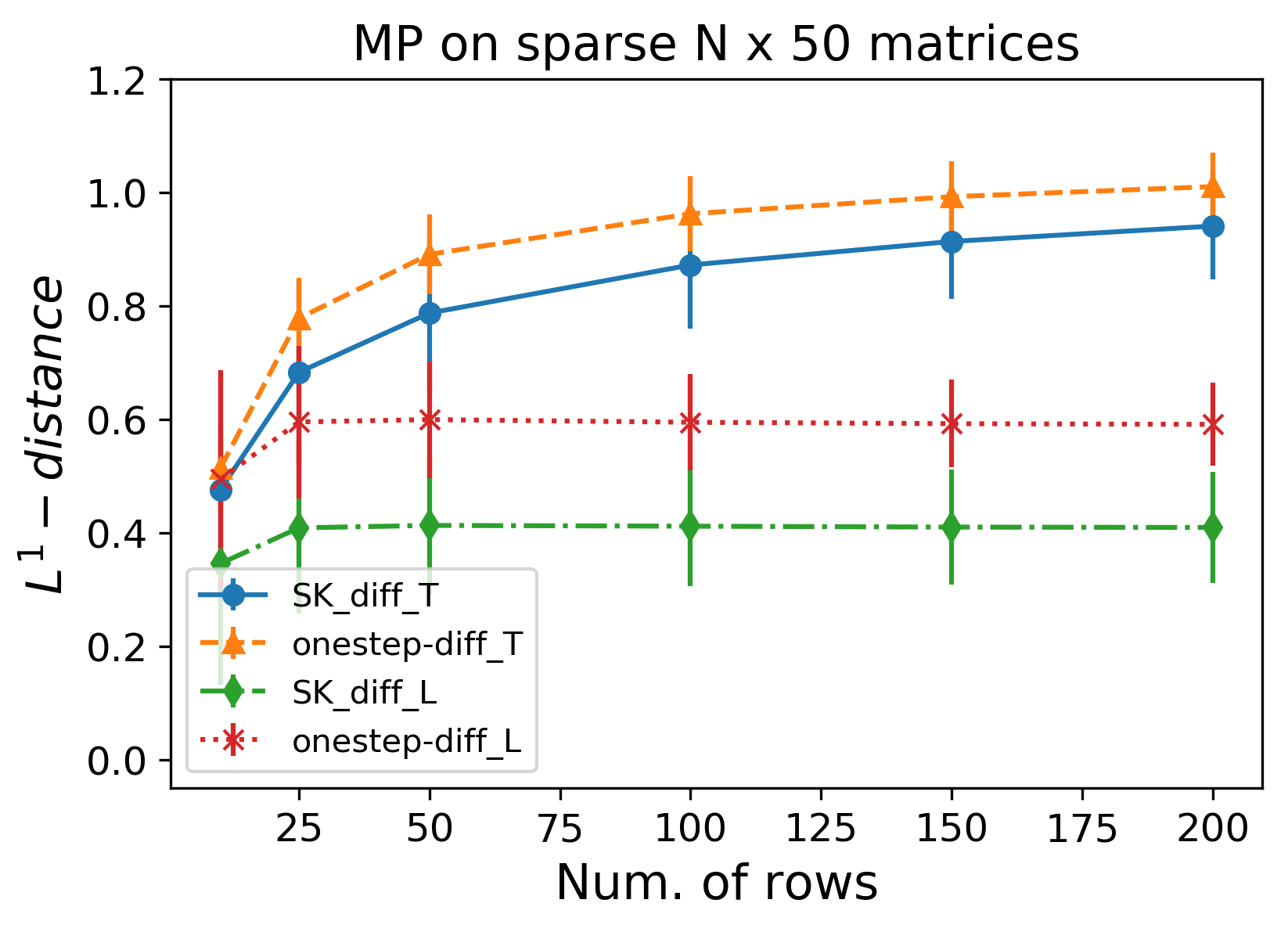}
    \caption{\textbf{a.} The Cooperative Index (CI) of Sinkhorn plans (SK) and its one step approximation (onestep) as total perturbation increases. 
    \textbf{b-d.} $r = 0.03$, $\epsilon =1$, dimension of $M$ varies as shown in $x$-axis. \textbf{b.} The probability of SK has higher CI than onestep. \textbf{c.} The average communication effectiveness for SK and onestep with and without perturbations denoted by SK-p, onestep-p, SK-np, onestep-np accordingly. \textbf{d.} The average difference of the teaching (and learning) plan for SK (and one-step approximation) before and after perturbations, measured by the $L^1$-distance.}
    \label{fig:rec}
\end{figure}

\textbf{Rectangular matrices.} Figure~\ref{fig:rec} are plots based on stimulation of matrix perturbation on rectangular matrices. 
The number of columns for sampled matrices is fixed to be  $50$.
The number of rows varies as in $[10, 25, 50, 100, 150, 200]$.
All the other parameters are the same as in the main text: $r = 0.03$, $\epsilon =1$ 
and parameter of Dirichlet distribution is $0.1$ for both initial matrix $M$ and prior over $\conceptSpace$.

\begin{figure}[h!]
    \centering
    \textbf{a.}\includegraphics[scale=0.4]{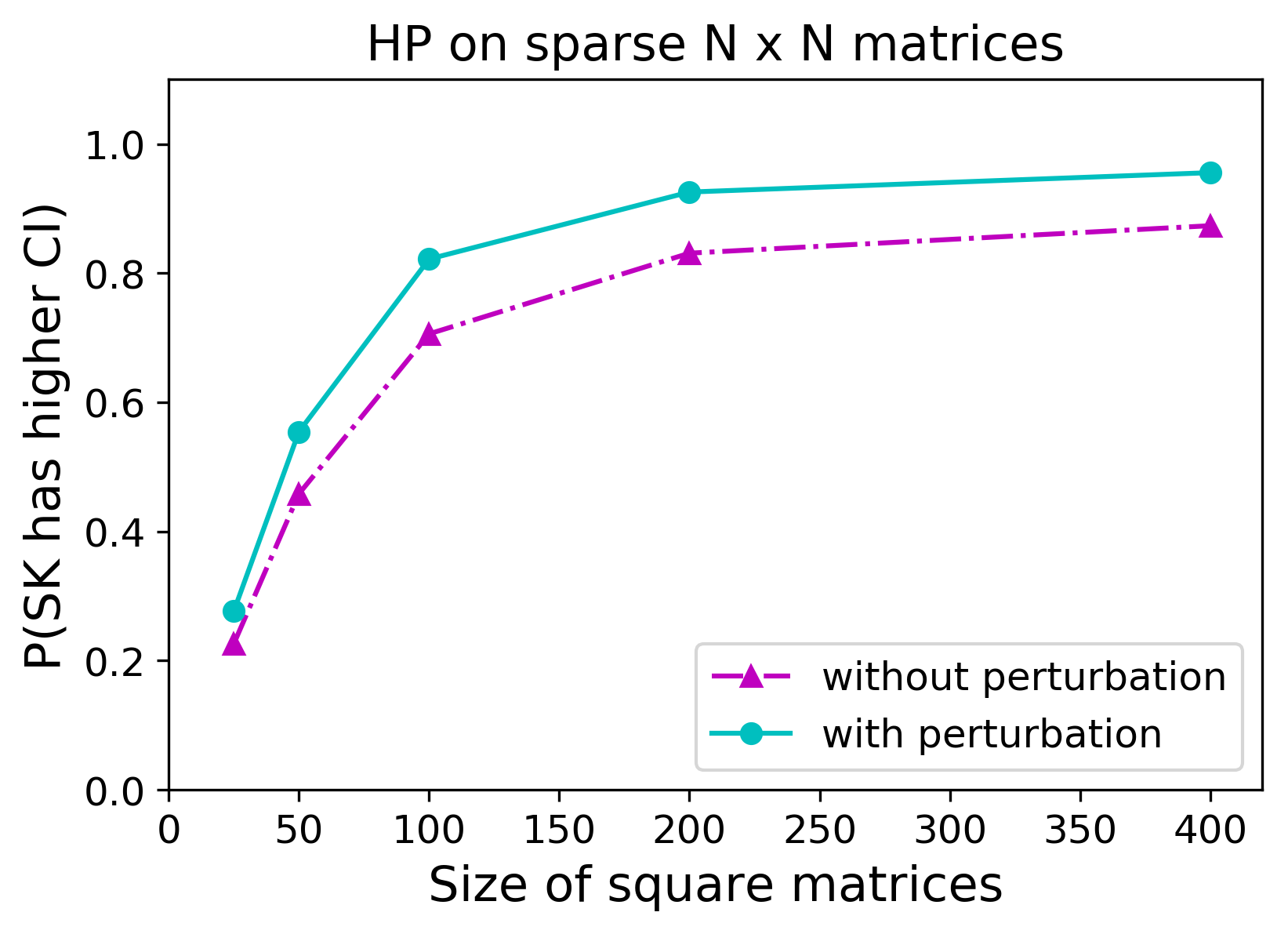}
    \textbf{b.}\includegraphics[scale=0.4]{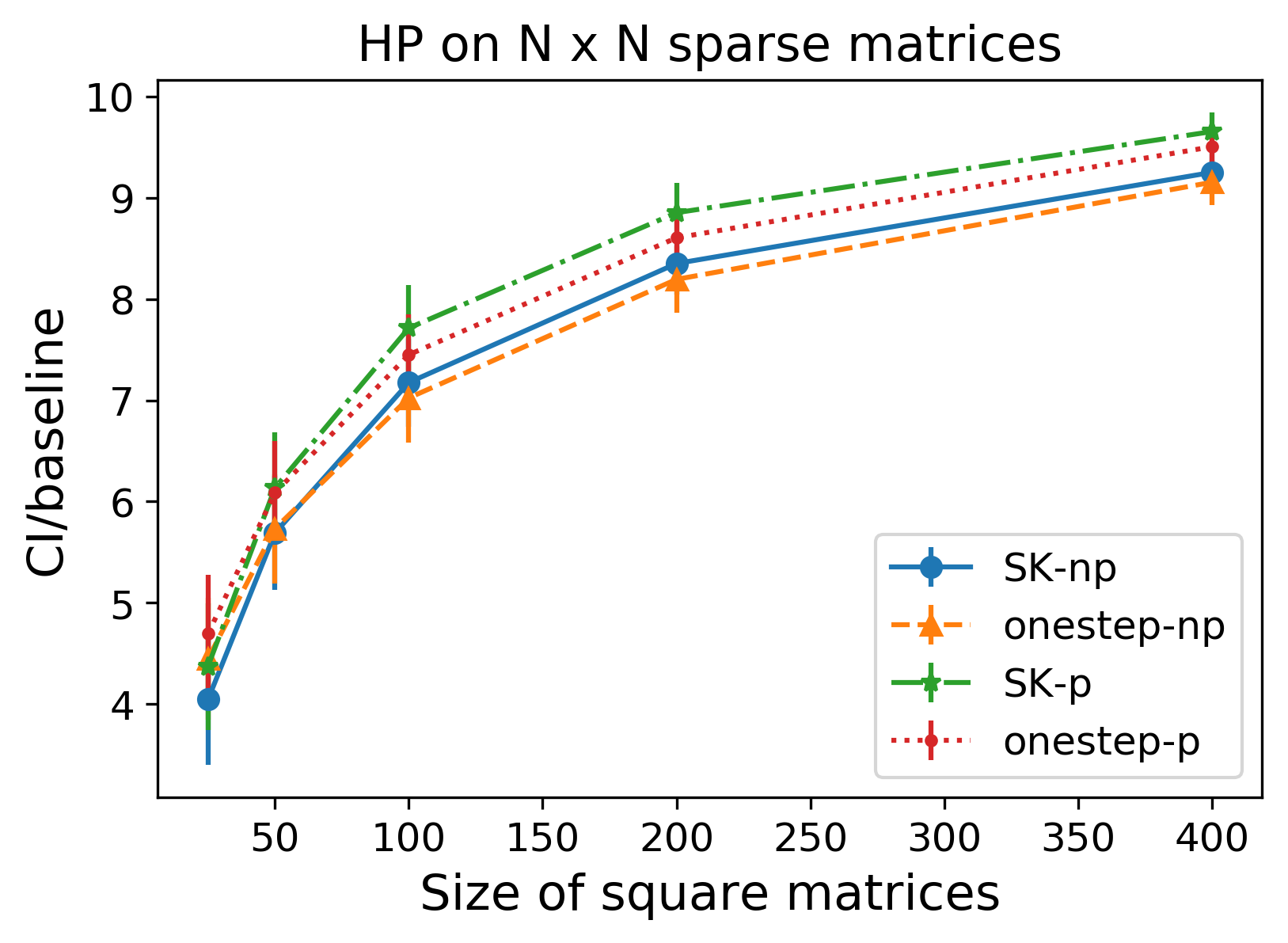}
    \textbf{c.}\includegraphics[scale=0.4]{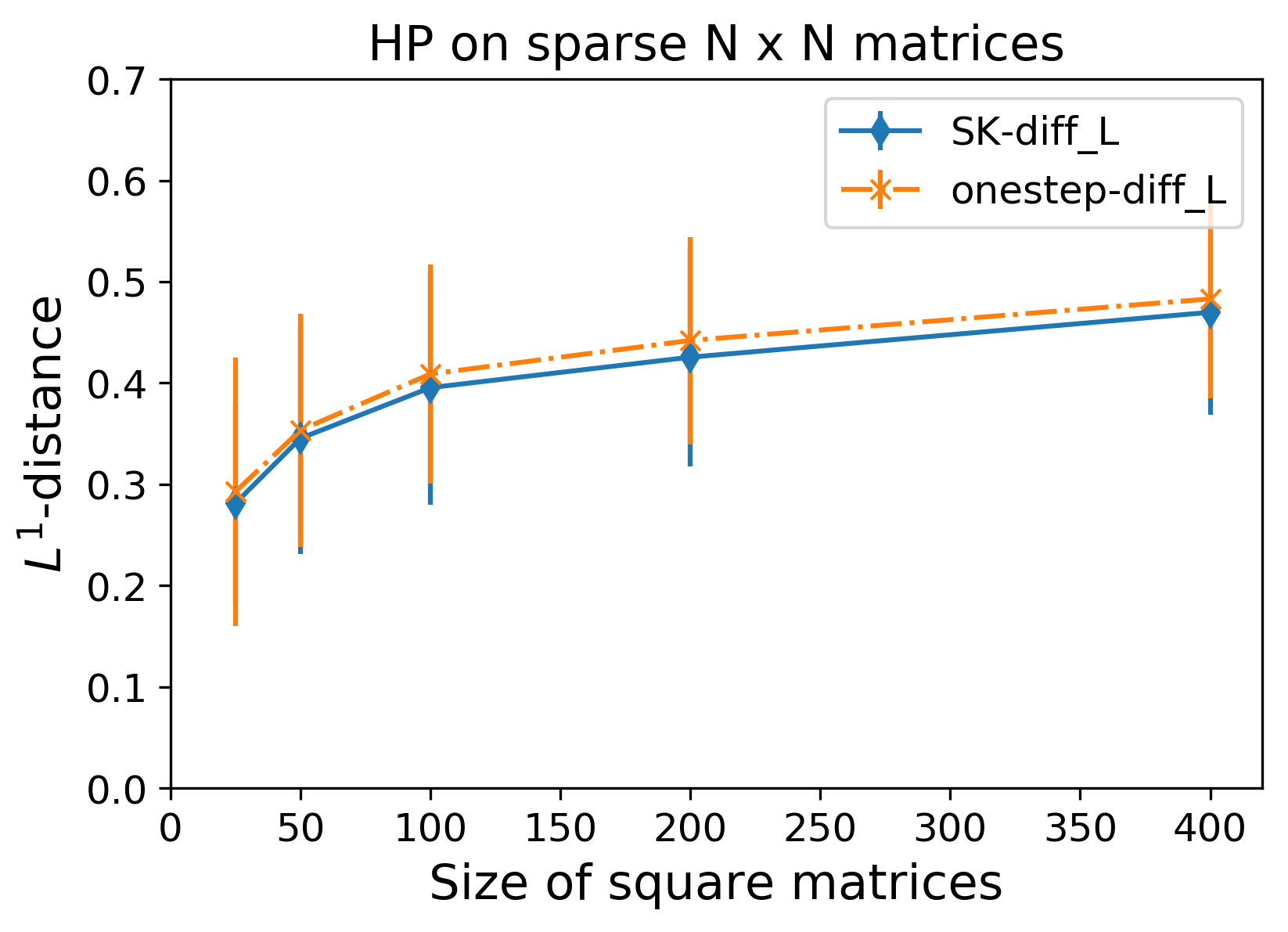}
    \textbf{d.}\includegraphics[scale=0.4]{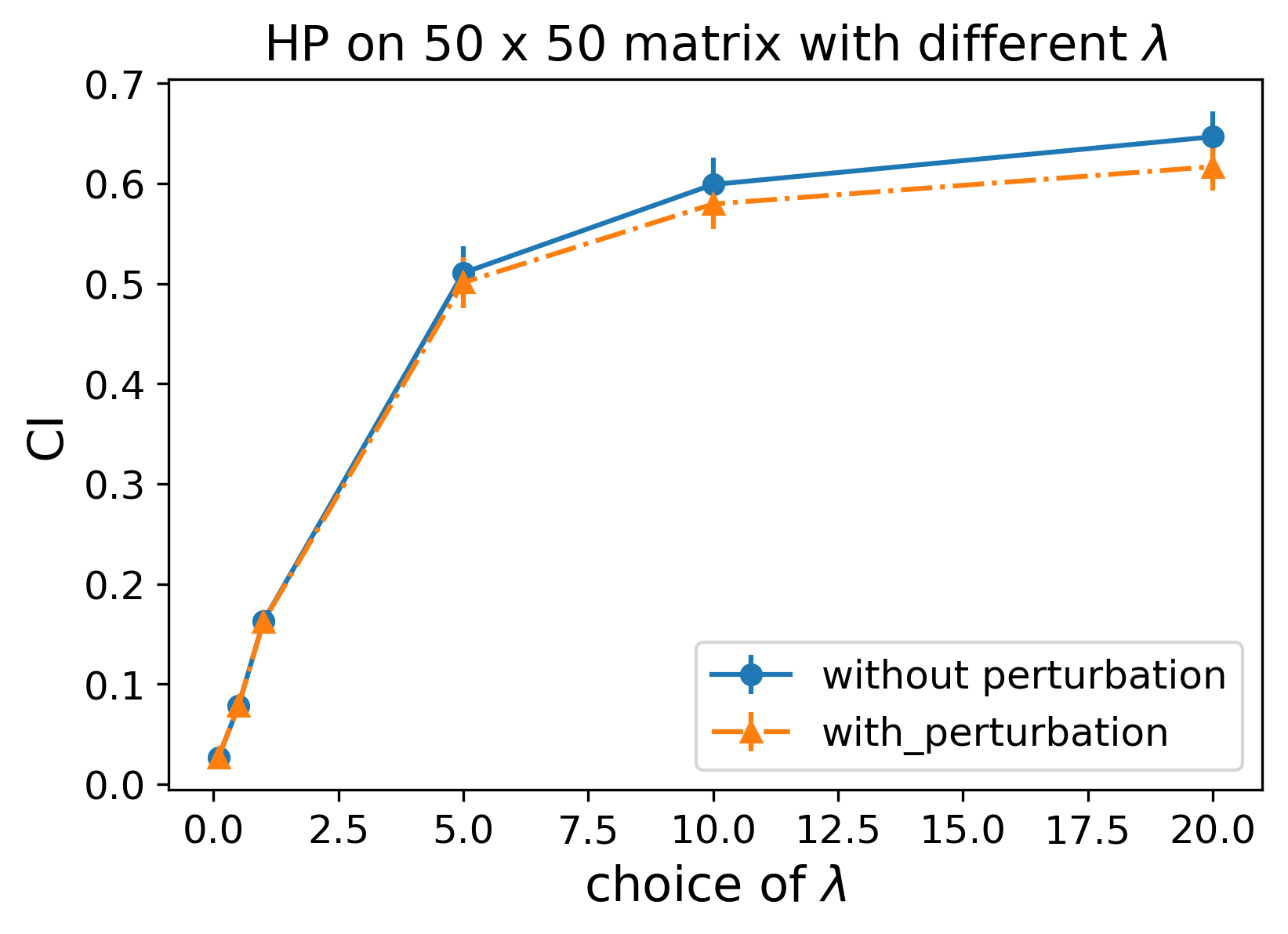}
    \caption{\textbf{a-c.} $r=0.07$, $\epsilon = 0.15$, dimension of $M$ varies as shown in $x$-axis. \textbf{a.}The probability of SK has higher CI than onestep. \textbf{b.}~The average communication efficiency for SK and onestep with and without perturbations denoted by SK-p, onestep-p, SK-np, onestep-np accordingly. \textbf{c.}~The average difference of the learning plan for SK and one-step approximation before and after perturbations, measured by the $L^1$-distance. \textbf{d.} The average CI for $50\times 50$ matrices as $\lambda$ varies in $[0.1,0.5,1,5,10,20]$.}
    \label{fig:hp}
\end{figure}

\textbf{Prior perturbation.} Figure~\ref{fig:hp} are plots based on stimulation of prior perturbation on square matrices. 
In \textbf{(a-c)}, the matrix size varies as in $[25, 50, 100, 200,400]$, 
parameter of Dirichlet distribution is $0.1$ for both initial matrix $M$ and prior over $\conceptSpace$.
We increase the perturbation rate to $r=0.07$ and reduce the magnitude to $\epsilon = 0.15$ 
as the prior over $\conceptSpace$ contains considerably fewer number of elements than $M$.
In \textbf{(d)}, the matrix size is fixed to be $50\times 50$, parameter of Dirichlet distribution for initial matrix is $10$,
for prior is $1$, $r=0.3$ and $\epsilon=0.3$. 
In general, we observer that both Sinkhorn plans and its one step approximation are much more sensitive to matrix perturbations than prior perturbations.
    
\subsection{Linear Approximations} \label{apd:sub:LA}
\begin{figure}[h!]
    \centering
    \textbf{a.}\includegraphics[scale=0.45]{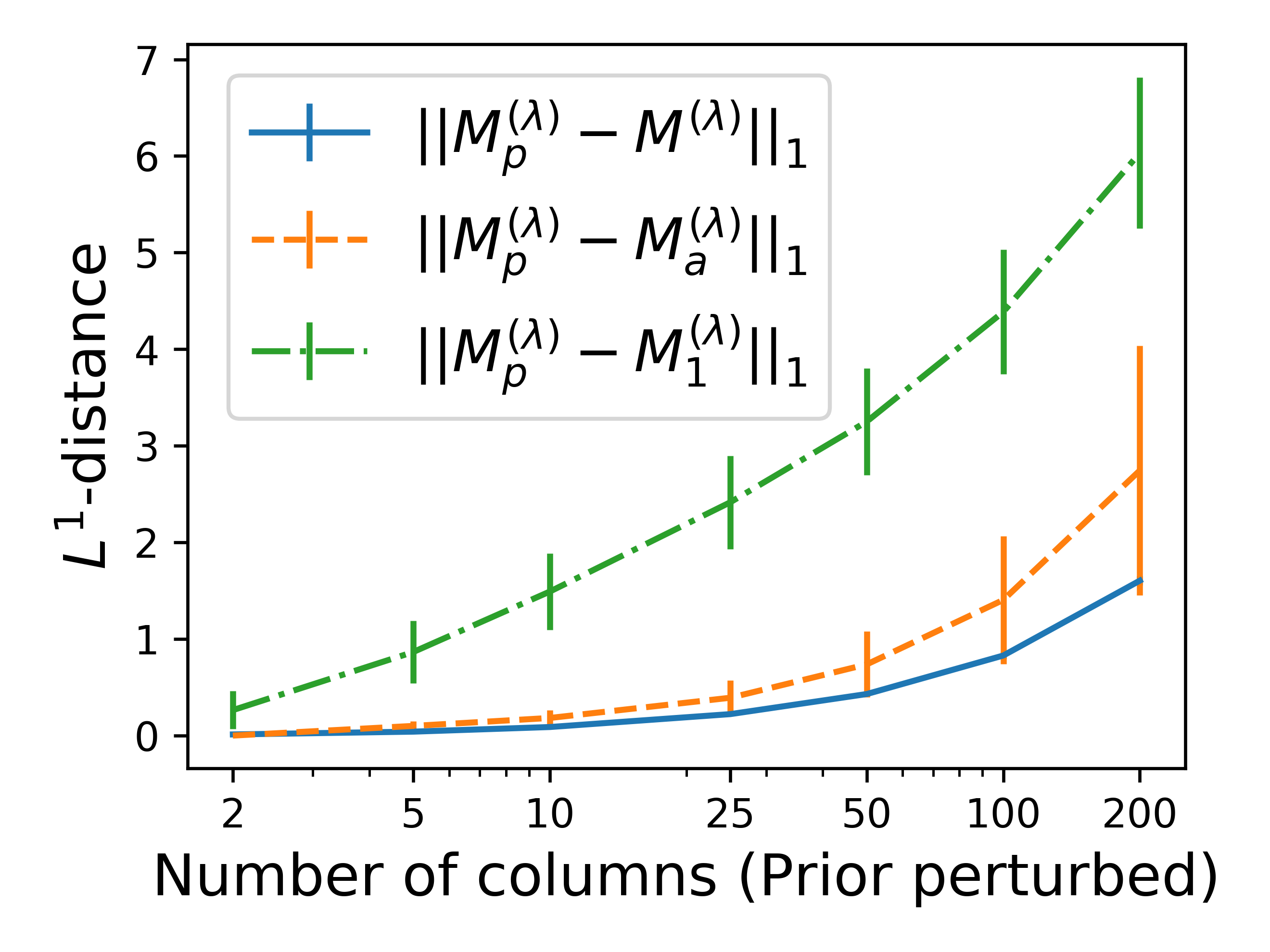}
    \textbf{b.}\includegraphics[scale=0.45]{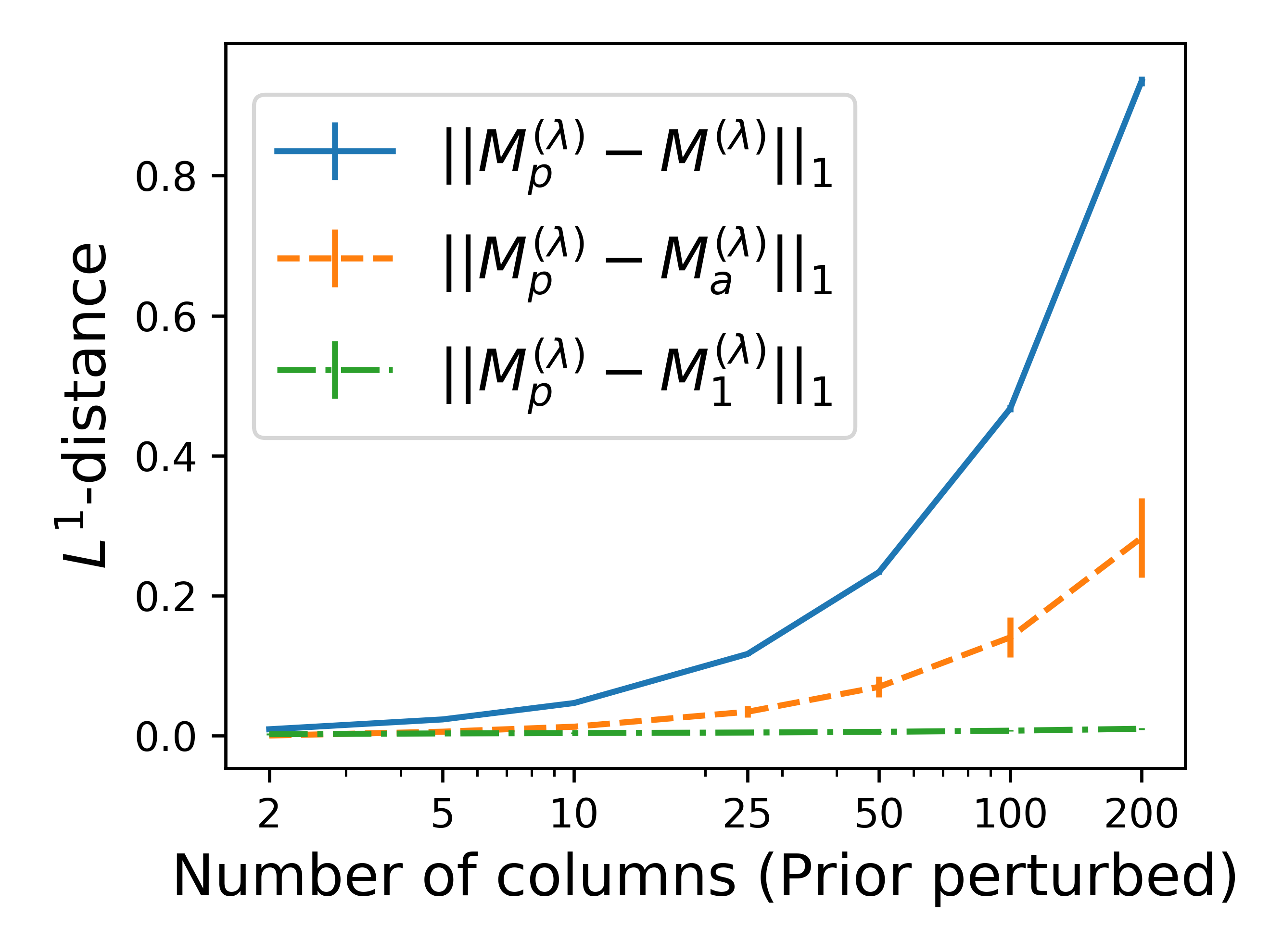}
    \textbf{c.}\includegraphics[scale=0.45]{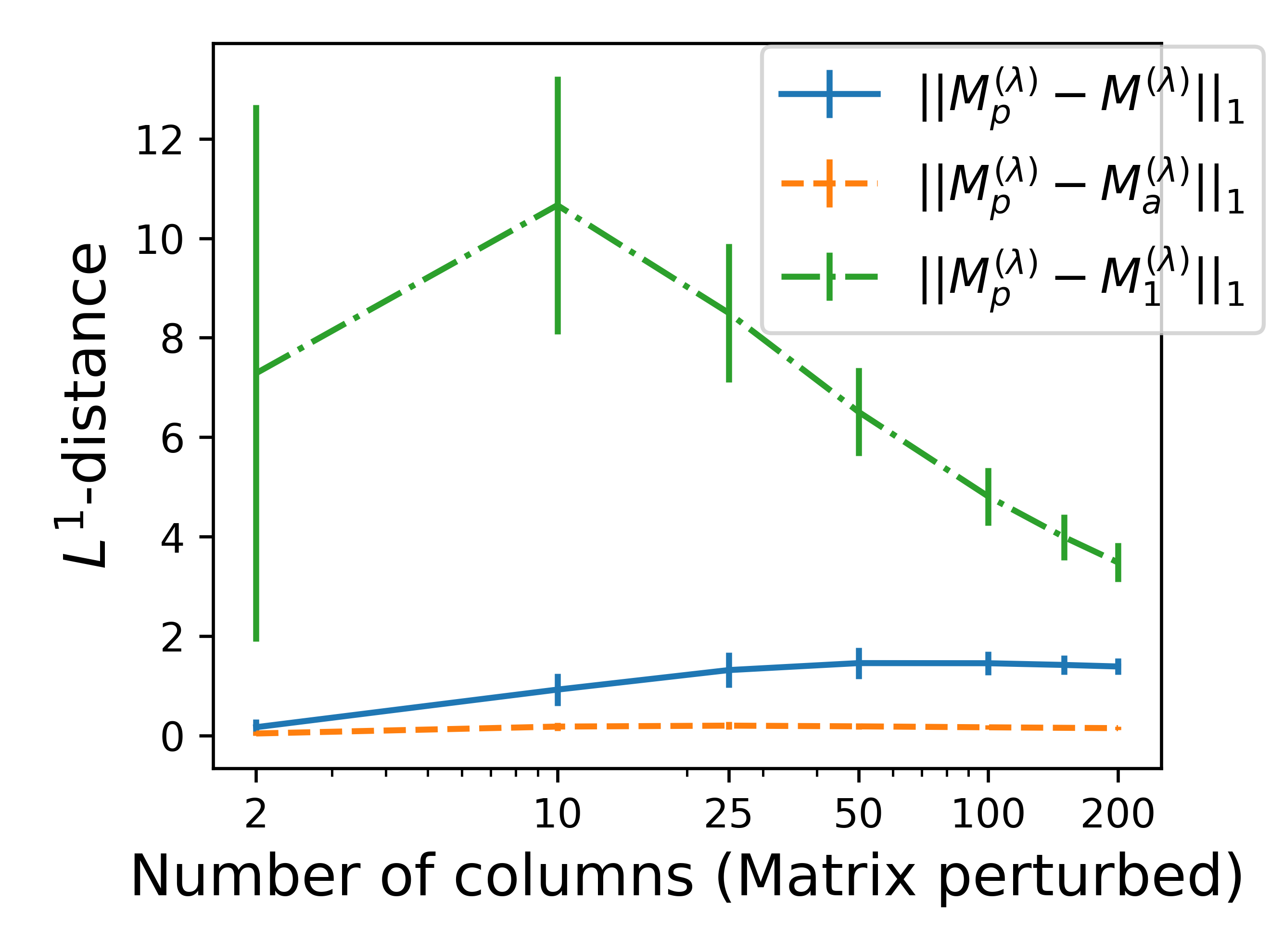}
    \textbf{d.}\includegraphics[scale=0.45]{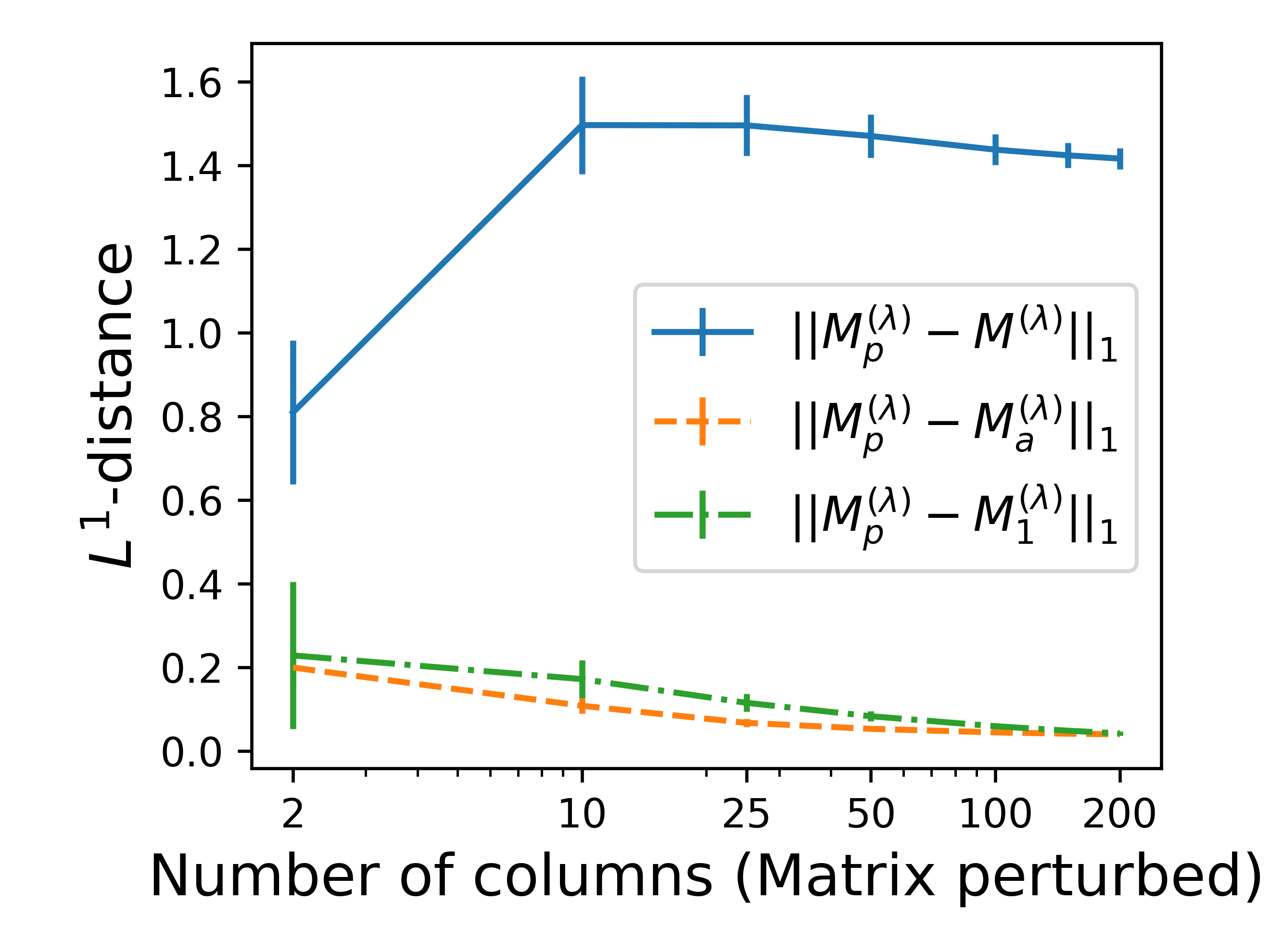}
    \caption{\textbf{a.} Perturb on row sums, with $\alpha=0.1$;
    \textbf{b.} Perturb on row sums, with $\alpha=10$;
    \textbf{c.} Perturb on matrices, with $\alpha=0.1$;
    \textbf{d.} Perturb on matrices, with $\alpha=10$.}
    \label{fig:linear-supp}
\end{figure}

Figure~\ref{fig:linear-supp} shows the result of comparisons on
different approximations of Sinkhorn limits of perturbed 
matrices/marginals, with different choices of Dirichlet hyperparameter
$\alpha=0.1,10$ ($\alpha=1$ in the main paper). Other
parameters (matrix size, sample size and method, and 
perturbation patterns) are the same as in the main text.

\section{Proofs of Propositions} \label{apd:eg_proof}

\begin{repprop}{prop:ci_is_ot}
Optimal conditional communication plans, $T^{\star}$ and $L^{\star}$, 
of a cooperative inference with arbitrary priors denoted by $P_{T_{0}}(\dataSpace)$ and $P_{L_{0}}(\conceptSpace)$, can be obtained through Sinkhorn scaling. In particular, as a direct consequence, cooperative inference is a special case of the unifying EOT framework with $\lambda=1$.
\end{repprop}


\begin{proof}

Consider cooperative inference as in Eq.~\eqref{eq:CI} of the main content, we may rewrite it as follows:
\begin{eqnarray}
 \label{eq:CI-var}
  P_{L}(h|d)P_{T_0}(d) &=& \frac{P_{T}(d|h) P_{L_{0}}(h)P_{T_0}(d)}{P_{L}(d)}\nonumber\\
 P_{T}(d|h)P_{L_0}(h) &=& \frac{P_{L}(h|d) P_{T_{0}}(d)P_{L_0}(h)}{P_T(h)}
\end{eqnarray}
which is equivalent to 

\begin{subequations}
\begin{align}
P_{L}(h|d)P_{T_0}(d) &= \frac{P_{T}(d|h) P_{L_{0}}(h)}{P_{L}(d)/P_{T_0}(d)},
\label{eq:L-var2}\\
P_{T}(d|h)P_{L_0}(h) &= \frac{P_{L}(h|d) P_{T_{0}}(d)}{P_T(h)/P_{L_0}(h)}.
\label{eq:T-var2}
\end{align}
 \label{eq:CI-var2}
\end{subequations}
\noindent Notice that Eq.~\eqref{eq:CI-var2} is the stable condition of Sinkhorn scaling 
on $\widetilde{M}=P_{L}(h|d)P_{T_0}(d)$ with $\mathbf{r}=P_{T_0} (\dataSpace)$, $\mathbf{c}=P_{L_0}(\conceptSpace)$.
Hence Eq.~\eqref{eq:CI-var2} can be solved using fixed-point iteration as explored in \citep{Shafto2014}: 
for the first evaluation of the left hand side of \eqref{eq:L-var2}, initialize $P_{L}(h|d)$ by $P_{L_0}(h|d)$ which is the row normalization of the shared distribution $M=P(d,h)$ and denote $P_{L_0}(h|d)\cdot P_{T_0}(d)$ by $\widetilde{L}_0$. 
Then the first evaluation of the left hand side of \eqref{eq:T-var2}, denoted by $\widetilde{T_1}$, can be obtained by column normalizing $\widetilde{L}_0$ with respect to $\cc$. Next, the second evaluation of \eqref{eq:L-var2} is achieved by row normalizing 
of $\widetilde{T_1}$ with respect to $\rr$, and iterate this process until convergence. This is precisely $(\rr, \cc)$-Sinkhorn scaling
starting with $\widetilde{L}_0$. Symmetrically, \eqref{eq:CI-var2} can also be solved by $(\rr, \cc)$-Sinkhorn scaling starting with 
$\widetilde{T}_0 = P_{T_0}(d|h)\cdot P_{L_0}(h)$.

Let $\M$ be the shared distribution, 
$\rr= P_{T_0}(\dataSpace)$ be the teacher's prior and $\cc=P_{L_0}(\conceptSpace)$ be learner's prior.
As shown in the above paragraph, after cooperative inference, 
the teacher’s conditional communication plan $T^{\star}$ is the limit
of $(\cc, \rr)$-SK scaling of $\widetilde{L}_0 =(P_{L_0}(h_j|d_i)P_{T_0}(d_i))$.
On the other hand, under the unifying EOT framework,  
the optimal teaching plan $T^{(\lambda=1)}$ is the limit of $(\cc,\rr)$-SK scaling of $\widehat{L}_0 =(P_{L_0}(h_j|d_i)e^{S_T(d_i)})$ based on Eq.~\eqref{eq:ot_teaching}.
When the teacher’s expense $S_T(d_i)$ of selecting $d_i$ is proportional to $\log P_{T_0}(d_i)$, $T^{(1)} = T^{\star}$.
Symmetrically, one may check the same holds for $L^{(1)} = L^{\star}$.

\end{proof}

\begin{repprop}{prop: lambda}
Assuming uniform marginals, $M^{(\lambda)}$ is concentrating around the leading diagonals of $M$ as $\lambda \to \infty$.
\end{repprop}

\begin{proof}
Let $D_{\sigma}, D_{\sigma'}$ be two diagonals of a $n\times n$ shared matrix $M$ and $d_{\sigma},d_{\sigma'}$ 
be products of their elements respectively (Definition~\ref{def:diag}). 
Further, let the diagonals in $M^{[\lambda]}$ determined by 
the same $\sigma$ and $\sigma'$ be $D^{[\lambda]}_{\sigma}$ and $D^{[\lambda]}_{\sigma'}$. 
Their cross product ratio is denoted by $\CR(D^{[\lambda]}_{\sigma}, D^{[\lambda]}_{\sigma'})$.
If $D_{\sigma'}$ is a leading diagonal and $D_{\sigma}$ is not, then $d_{\sigma}/d_{\sigma'}<1$,  and so
$\CR(D^{[\lambda]}_{\sigma}, D^{[\lambda]}_{\sigma'}) = (d_{\sigma}/d_{\sigma'})^{\lambda} \to 0$ as $\lambda\to \infty$ (Fact~$A$).
If both $D_{\sigma}$ and $D_{\sigma'}$ are leading diagonals, then  $d_{\sigma}/d_{\sigma'}=1$, and so
$\CR(D^{[\lambda]}_{\sigma}, D^{[\lambda]}_{\sigma'}) = (d_{\sigma}/d_{\sigma'})^{\lambda} \to 1$ as $\lambda \to \infty$.
We now show that for any element $M^{(\lambda)}_{st}$ of $M^{(\lambda)}$, if the corresponding element $M_{st}$ is not on a 
leading diagonal of $M$, then $M^{(\lambda)}_{st}\to 0$.
It is clear that if $M_{st}$ is not contained in any positive diagonal of $M$, then $M^{(\lambda)}_{st}\to 0$
as off diagonal elements vanishes along Sinkhorn iteration \citep{wang2018generalizing}.
Now suppose that $M_{st}$ is contained in a non-leading positive diagonal determined by permutation $\sigma$.
If $M^{(\lambda)}_{st}$ does not vanish, there exists an $\e>0$ such that $M^{(\lambda)}_{st}>\e$ for any $\lambda$.
And so $M^{(\lambda)}_{st}$ must be contained in a positive diagonal of $M^{(\lambda)}$.
Without loss, we may assume $M^{(\lambda)}_{st}$ is the smallest non-vanishing element that is off leading diagonals of $M$.
Then $d^{(\lambda)}_{\sigma}>\e^n$, and so $d^{(\lambda)}_{\sigma}/d^{(\lambda)}_{\sigma'}>\e^n$  because $d^{(\lambda)}_{\sigma'} \leq 1$ 
($M^{(\lambda)}$ is a joint distribution). 
This is contradiction to Fact~$A$.
Therefore, $M^{(\lambda)}$ is concentrating around the leading diagonals of $M$ as $\lambda \to \infty$. 
\end{proof}


\cite{wang2018generalizing} explored the sensitivity of $\Phi$ to perturbation on elements in $M$. They showed that $\Phi$ is continuous on $M$. In particular, they demonstrated that $\Phi$ is robust to any amount of off-diagonal perturbations on $M$. SK scaling is also continuous on its scalars. Let $\rr^{\e}$ and $\cc^\e$ be vectors obtained by varying elements of $\rr$ and $\cc$ at most by $\e$, where $\e>0$ quantifies the amount of perturbation. Distances between vectors or matrices are measured by $l^{\infty}$ norm (the maximum element-wise difference), e.g. $d(\rr^{\e}, \rr)\leq \e$. 
We prove that $\Phi$ is continuous on $\rr$ and $\cc$, thus the following holds:

\begin{repprop}{thm: sk_cont}
For any joint distribution $M$ and positive marginals $\rr$ and $\cc$,
if $\Phi(M, \rr^\e, \cc^\e)$ and $ \Phi(M,\rr,\cc)$ exist, then
$\Phi(M, \rr^\e, \cc^\e)\to \Phi(M,\rr,\cc)$ as $\rr^\e \to \rr, \cc^\e \to \cc$.
\end{repprop}

\begin{proof}
Note that the continuity of $\Phi$ on the marginals is independent of the choice of a particular $\lambda$,
we will drop the $\lambda$ for the rest of the proof to make the notation neater.
Sinkhorn scaling of $\M$ converges with marginal conditions $(\rr,\cc)$ and $(\rr^{\e}, \cc^{\e})$ implies that
$\sum_{i=1}^{n} r_i = \sum_{j=1}^{m}c_j$ and $ \sum_{i=1}^{n} r^{\e}_i = \sum_{j=1}^{m}c^{\e}_j$ (see \cite{menon1969spectrum}).
Let $k = \sum_{i=1}^{n} r_i $ and $k^{\e} =  \sum_{i=1}^{n} r^{\e}_i$.
We will prove in three steps. 
First, we show the claim when $k=k^{\e}$.
As $k=k^{\e}$, at least two elements in $\rr$ (or $\cc$) are perturbed.
Without loss, we will assume that only two elements, $r_s$ and $r_t$ in $\rr$,  are varied by amount $\e$ since 
the general case may be treated as compositions of such.
Then for $\rr^{\e} = (r^{\e}_1, \dots, r^{\e}_n)$, we have
$r^{\e}_s = r_s +\e$, $r^{\e}_t = r_t -\e$ and $r^{\e}_i = r_i$ if $i\neq s \text{ or } t$. 
Let $\Phi(M, \rr, \cc) = M^*$, $M^{*\e}$ be the matrix obtained from 
varying the element $M^{*}_{s1}$ and $M^{*}_{t1}$ of $M^*$ by $\e$ and $-\e$,
i.e. $M^{*\e}_{s1} = M^*_{s1}+\e$, $M^{*\e}_{t1} = M^*_{t1}-\e$ and $M^{*\e}_{ij} = M^*_{ij}$ otherwise.
Then the statement can be verified as following:
\begin{align*} 
d(\Phi(&M, \rr, \cc), \Phi(M, \rr^{\e}, \cc)) \overset{(a)}{=}d(M^*, \Phi(M^*, \rr^{\e}, \cc))\\
& \overset{(b)}{\leq}  d(M^*, \Phi(M^{*\e}, \rr^\e, \cc)) + d(\Phi(M^{*\e}, \rr^\e, \cc), \Phi(M^*, \rr^{\e}, \cc))\\
& \overset{(c)}{=} d(M^*, M^{*\e}) + d(\Phi(M^{*\e}, \rr^\e, \cc), \Phi(M^*, \rr^{\e}, \cc))\\
& \overset{(d)}{=} \e + d(\Phi(M^{*\e}, \rr^\e, \cc), \Phi(M^*, \rr^{\e}, \cc))\overset{(e)}{\to} 0 \text{ as } \e \to 0
\end{align*}
where $(a)$ holds since $M^*$ and $M$ are cross-ratio equivalent and must converge to the same limit under any Sinkhorn scaling; 
$(b)$ is triangle inequality; $(c)$ holds since $M^{*\e}$ is already $(\rr^{\e},\cc)$-normalized, hence
$\Phi(M^{*\e}, \rr^{\e},\cc) = M^{*\e}$;
$(d)$ holds as $d(M^*, M^{*\e}) = \e$ by construction; $(e)$ holds because $\Phi$ is continuous on $M$ proved in \cite{sinkhorn1972continuous}.


Now we show the case where $k \neq k^{\e}$, but the proportion between corresponding elements in $\rr$ and $\rr^{\e}$ are the same, thus $r^{\e}_i/r_i = r^{\e}_j/r_j =\alpha$ . Let $M^{*\alpha} = \alpha * M^*$, i.e. $M^{*\alpha}_{ij} = \alpha * M^*_{ij}$.
Since $M^{*\alpha}$ is $(\rr^{\e}, \cc)$ normalized and also has the same cross ratios of $\M$, $\Phi(M, \rr^{\e}, \cc) = M^{*\alpha}$.
Note that $d(M^{*\alpha}, M^*) \leq \e$, so $\Phi(M, \rr^{\e}, \cc) \to \Phi(M, \rr, \cc)$ as $\e \to 0$.

Finally for the general case, where $k \neq k^{\e}$ and elements of $\rr$ and $\rr^{\e}$ are not proportional.
Let $\rr^{\alpha} = (k^{\e}/k) * \rr$. Then elements of $\rr$ and $\rr^{\alpha}$ are proportional and $\sum \rr^{\alpha}_{i} = \sum \rr^{\e}_{i} = k^{\e}$.
Thus based on the previous two cases, we have 
$d(\Phi(M, \rr, \cc), \Phi(M, \rr^{\e}, \cc)) \leq d(\Phi(M, \rr, \cc), \Phi(M, \rr^{\alpha}, \cc)) +
d(\Phi(M, \rr^{\alpha}, \cc), \Phi(M, \rr^{\e}, \cc)) \to 0$ as $\e\to 0$. Hence, we are done.
\end{proof}


\subsection{General version of Theorem~\ref{thm:smoothness}} \label{apd: smooth}

Enlightened by \cite{luise2018differential}, we can conclude a stronger version of the smoothness of $\Phi$ in the following way:

\begin{definition*}
A pattern $\mathfrak{P}$ is a subset
of $\{1,2,\dots,n\}\times\{1,2,\dots,m\}$, and a matrix $M=(M_{ij})$ of pattern $\mathfrak{P}$
is a non-negative matrix with $M_{ij}>0$ if and only if
$(i,j)\in\mathfrak{P}$. In this paper,
$M$ is not allowed to have a vanishing row or column.
\end{definition*}

\begin{repthm}{thm:smoothness} [General venison of Theorem~\ref{thm:smoothness}]
  Let $(\mathfrak{P},\mathfrak{D})$ be a pair where
  $\mathfrak{P}$ is a pattern, and where
  $\mathfrak{D}\subseteq(\mathbb{R}^+)^{n}\times(\mathbb{R}^+)^m$ is the set consisting of vectors
  $(\mathbf{r},\mathbf{c})\in(\mathbb{R}^+)^n\times(\mathbb{R}^+)^m$ satisfying the equivalent conditions in Theorem 2 of
  \cite{rothblum1989scalability}, in other words, pattern $\mathfrak{P}$ is exact
  $(\mathbf{r},\mathbf{c})$-scalable. Let
  $\mathcal{M}_{\mathfrak{P}}=(\mathbb{R}^+)^{\mathfrak{P}}$ be the open cone of
  nonnegative matrices of pattern $\mathfrak{P}$, then for a given
  $\lambda\in(0,\infty)$, 
  $\Phi:\mathcal{M}_{\mathfrak{P}}\times\mathfrak{D}\rightarrow\mathcal{M}_{\mathfrak{P}}$ is smooth.
 \end{repthm}
  
\begin{proof}
  We use the same strategy as the proof of Theorem 2 in \cite{luise2018differential}.
  Throughout the proof, let $\lambda\in(0,\infty)$ be a fixed positive real
  number. 

  First we make a decomposition of $\Phi$.
  This is possible because the exact scaling conditions guarantee the existence of
  diagonal matrices $D_1,D_2$ such that
  $\Phi(M,\mathbf{r},\mathbf{c})=M^{(\lambda)}=D_1M^{[\lambda]}D_2$,
  equivalently, there exist a pair of vectors
  $(\alpha,\beta)\in\mathbb{R}^n\times\mathbb{R}^m$ such that 
  $\Phi(M,\mathbf{r},\mathbf{c})=\mathrm{diag}(e^{\lambda\alpha})M^{[\lambda]}\mathrm{diag}(e^{\lambda\beta})$.
  The pair $(D_1,D_2)$ is unique up to a scalar $d\in\mathbb{R}^+$ with actions
  $d:(D_1,D_2)\mapsto(dD_1,d^{-1}D_2)$, thus the pair of vectors
  $(\alpha,\beta)$ is unique up to a constant
  $\delta:(\alpha,\beta)\mapsto(\alpha+\delta,\beta-\delta)$ (plus/minus the
  same number on each element of the vectors). So we may always assume that the
  last component of $\beta$ vanishes, i.e., $\beta_m=0$. In the following text,
  we use $\bar{\beta}$ to denote the first $m-1$ components of $\beta$, and
  if $\bar{\beta}$ occurs, the corresponding $\beta$ is the vector by appending a
  $0$ at the end of $\bar{\beta}$.

  Then we can decompose the map $\Phi$ into the composition of two other maps:
  $\Phi=\mu\circ(\rho,\Psi)$. Here the map
  $\rho:\mathcal{M}_{\mathfrak{P}}\times\mathfrak{D}\rightarrow\mathcal{M}_{\mathfrak{P}}$
  is the regularization map (regardless of the marginal conditions)
  $\rho(M,(\mathbf{r},\mathbf{c}))=M^{[\lambda]}$, the map
  $\Psi:\mathcal{M}_{\mathfrak{P}}\times\mathfrak{D}\rightarrow\mathbb{R}^n\times\mathbb{R}^m$
  maps $(M,\mathbf{r},\mathbf{c})$ to the pair of vectors $(\alpha,\beta)$ with
  $\beta_m=0$ as in the above discussion (such that
  $\Phi(M,\mathbf{r},\mathbf{c})=\mathrm{diag}(e^{\lambda\alpha})M^{[\lambda]}\mathrm{diag}(e^{\lambda\beta})$),
  and the map
  $\mu:\mathcal{M}_{\mathfrak{P}}\times\mathbb{R}^n\times\mathbb{R}^m\rightarrow\mathcal{M}_{\mathfrak{P}}$
  is such that
  $\mu(P,\alpha,\beta)=\mathrm{diag}(e^{\lambda\alpha})(P)\mathrm{diag}(e^{\lambda\beta})$.
  It can be easily seen that from the definitions the decomposition
  $\Phi=\mu\circ(\rho,\Psi)$ is valid.

  Next, having this decomposition, we just need to show that $\mu$, $\rho$ and $\Psi$
  are smooth, then $\Phi$ as the composition of smooth maps remains smooth.

  (Smoothness of $\Psi$:) We use the same strategy as Theorem 2 in
  \cite{luise2018differential}.
  Define the Lagrangian
  $$\mathcal{L}(M,\mathbf{r},\mathbf{c};\alpha,\beta)=
    -\mathbf{r}^\top\alpha-\mathbf{c}^\top\beta+
    \sum_{(i,j)\in\mathfrak{P}}\dfrac{e^{\lambda\alpha_i}M_{ij}^{\lambda}e^{\lambda\beta_j}}{\lambda}.$$
where  $\Psi(M,\mathbf{r},\mathbf{c})=(\alpha,\beta)$ optimizes $\mathcal{L}$
  for fixed $M$, $\mathbf{r}$, $\mathbf{c}$ as proved in \cite{luise2018differential,cuturi2013sinkhorn}.
  By smoothness of $\mathcal{L}$ (easy to see from expression), we may conclude
  that $N:=\nabla_{(\alpha,\bar{\beta})}\mathcal{L}$ is $C^k$ for any $k\ge0$ and
  $\nabla_{(\alpha,\bar{\beta})}\mathcal{L}(M,\mathbf{r},\mathbf{c};\Psi(M,\mathbf{r},\mathbf{c}))=\mathbf{0}$
  for any $M,\mathbf{r},\mathbf{c}$.

  Fix $(M_0,\mathbf{r}_0,\mathbf{c}_0;\alpha_0,\beta_0)$ such that
  $N(M_0,\mathbf{r}_0,\mathbf{c}_0;\alpha_0,\beta_0)=\mathbf{0}$ and $(\beta_0)_m=0$. Since
  $\nabla_{(\alpha,\bar{\beta})}N=\nabla_{(\alpha,\bar{\beta})}\otimes\nabla_{(\alpha,\bar{\beta})}\mathcal{L}$
  is the Hessian of the strictly convex function $\mathcal{L}$, then
  $\nabla_{(\alpha,\bar{\beta})}N(M_0,\mathbf{r}_0,\mathbf{c}_0;\alpha_0,\beta_0)$ is invertible.
  Thus by Implicit Function Theorem, there exists a neighbourhood $U$ of
  $(M_0,\mathbf{r}_0,\mathbf{c}_0)$ in
  $\mathcal{M}_{\mathfrak{P}}\times\mathfrak{D}$ and a map
  $\psi:U\rightarrow\mathbb{R}^n\times\mathbb{R}^m$ such that
  \begin{enumerate}
  \item
    $\psi(M_0,\mathbf{r}_0,\mathbf{c}_0)=(\alpha_0,\beta_0)$,
  \item denote $\psi(M,\mathbf{r},\mathbf{c})=(\alpha,\beta)$, then the last
    component of $\beta$ vanishes, $\beta_m=0$, for any
    $(M,\mathbf{r},\mathbf{c})\in U$,
  \item
    $N((M_0,\mathbf{r}_0,\mathbf{c}_0;\psi(M_0,\mathbf{r}_0,\mathbf{c}_0))=\mathbf{0}$,
    thus $\psi(M,\mathbf{r},\mathbf{c})=\Psi(M,\mathbf{r},\mathbf{c})$, $\forall (M,\mathbf{r},\mathbf{c})\in U$,
    by strict convexity of $\mathcal{L}$ and uniqueness of $(\alpha,\beta)$,
  \item $\psi\in C^k(U)$.
  \end{enumerate}

  For the choice of $k$ is arbitrary and the choice of
  $(M,\mathbf{r},\mathbf{c})$ as an interior point of
  $\mathcal{M}_{\mathfrak{P}}\times\mathfrak{D}$ is also arbitrary,
  we may see that $\Psi$ is smooth in the interior of
  $\mathcal{M}_{\mathfrak{P}}\times\mathfrak{D}$.

  In fact, we can show that
  $(\mathcal{M}_{\mathfrak{P}}\times\mathfrak{D})^\circ=\mathcal{M}_{\mathfrak{P}}\times\mathfrak{D}$,
  thus $\Psi$ is smooth on $\mathcal{M}_{\mathfrak{P}}\times\mathfrak{D}$.
  
  $\mathcal{M}_{\mathfrak{P}}$ is isomorphic to an open subset
  $\left(\mathbb{R}^+\right)^{|\mathfrak{P}|}$ of $\mathbb{R}^{|\mathfrak{P}|}$.
  The set $\mathfrak{D}$ is a subset of $\left(\mathbb{R}^+\right)^{n+m}$,
  defined by finitely many equations and strict inequalities given in
  \cite[Theorem 2]{rothblum1989scalability}, especially part (e):
  for every subset $I\subseteq\{1,2,\dots,n\}$ and $J\subseteq\{1,2,\dots,m\}$,
  where $M_{ij}=0$ for all $(i,j)\in I^c\times J$ ($I^c$ is the complement of
  $I$), we have
  $$\sum_{i\in I}r_i\ge\sum_{j\in J}c_j$$
  with equality holds if and only if $M_{ij}=0$ for all $(i,j)\in I\times J^c$.
  The above condition means that the conditions are either equations or strict
  inequalities since the pattern $\mathfrak{P}$ is fixed. Among all these constraints,
  set of equations $\mathcal{E}$ define a linear subspace $V(\mathcal{E})$ of
  $\mathbb{R}^{n+m}$ and the set of strict inequalities $\mathcal{N}$ draws an
  open subset $U(\mathcal{E},\mathcal{N})$ on $V(\mathcal{E})$. And
  $\mathfrak{D}=\left(\mathbb{R}^+\right)^{n+m}\cap U(\mathcal{E},\mathcal{N})$
  is open in $U(\mathcal{E},\mathcal{N})$,
  so $(\mathfrak{D})^\circ=\mathfrak{D}$.

  (Smoothness of $\rho$:) Since $\lambda>0$ and for each $(i,j)\in\mathfrak{P}$,
  $M_{ij}>0$, then $\rho$ is smooth from the smoothness of $x^\lambda$ on
  $(0,\infty)$.

  (Smoothness of $\mu$:) $\mu$ is the composition of exponential functions,
  multiplications and additions, all of which are smooth.

  Thus $\Phi=\mu\circ(\rho,\Psi)$ is smooth on
  $\mathcal{M}_{\mathfrak{P}}\times\mathfrak{D}$.

\subsection{Calculation of gradient of $\Phi$} \label{apd:gradient}
We make use of the decomposition $\Phi=\mu\circ(\rho,\Psi)$ to calculate the
gradient of $\Phi$.
  
  By implicit function theorem,
  \begin{eqnarray}
    \left(\nabla_{\mathbf{r}}\Psi\right)_i&=&\dfrac{\partial \Psi}{\partial\mathbf{r}_i}\nonumber\\
    &=&
    -\left(\nabla_{(\alpha,\bar{\beta})}N\right)^{-1}\left(\nabla_{\mathbf{r}}N\right)_i
                                          \nonumber\\
    &=&-\left(\nabla^2_{(\alpha,\bar{\beta})}\mathcal{L}\right)^{-1}\left(\nabla_{\mathbf{r}}N\right)_i
        \nonumber\\
    &=&-\dfrac{1}{\lambda}\left(
        \begin{array}{cc}
          \mathrm{diag}(\mathbf{r}) & \overline{M^{(\lambda)}}\\
          \overline{M^{(\lambda)}}^\top & \mathrm{diag}(\mathbf{\bar{c}})\\
        \end{array}
    \right)^{-1}\left(
    \begin{array}{c}
      (\mathbf{\delta_i})_n\\
      \mathbf{0}_{(m-1)}\\
    \end{array}
    \right)
    \nonumber\\
    &=&-\dfrac{1}{\lambda}\left(
        \begin{array}{cc}
          \mathrm{diag}(\mathbf{r}) & \overline{M^{(\lambda)}}\\
          \overline{M^{(\lambda)}}^\top & \mathrm{diag}(\mathbf{\bar{c}})\\
        \end{array}
    \right)^{-1}_{\text{col-}i}\nonumber
  \end{eqnarray}
  In the last equality, the subscript col-$i$ means the $i$-th column of the inverse
  matrix with $1\le i\le n$.


  \begin{eqnarray}
    \left(\nabla_M\Psi\right)_{ij}&=&\dfrac{\partial\Psi}{\partial
                                      M_{ij}}\nonumber\\
    &=&-\left(\nabla^2_{(\alpha,\bar{\beta})}\mathcal{L}\right)^{-1}\left(\nabla_{M}N\right)_{ij}
        \nonumber\\
    &=&\dfrac{1}{\lambda}\left(
        \begin{array}{cc}
          \mathrm{diag}(\mathbf{r}) & \overline{M^{(\lambda)}}\\
          \overline{M^{(\lambda)}}^\top & \mathrm{diag}(\mathbf{\bar{c}})\\
        \end{array}
    \right)^{-1}\cdot\lambda e^{\lambda(\alpha_i+\beta_j)}M_{ij}^{\lambda-1}\left(
    \begin{array}{c}
      \delta_i\\
      \bar{\delta_j}\\
    \end{array}
    \right)\nonumber\\
    &=&\dfrac{M^{(\lambda)}_{ij}}{M_{ij}}\left[\left(
        \begin{array}{cc}
          \mathrm{diag}(\mathbf{r}) & \overline{M^{(\lambda)}}\\
          \overline{M^{(\lambda)}}^\top & \mathrm{diag}(\mathbf{\bar{c}})\\
        \end{array}
    \right)^{-1}_{\text{col-}i}+\left(
        \begin{array}{cc}
          \mathrm{diag}(\mathbf{r}) & \overline{M^{(\lambda)}}\\
          \overline{M^{(\lambda)}}^\top & \mathrm{diag}(\mathbf{\bar{c}})\\
        \end{array}
    \right)^{-1}_{\text{col-}(n+\bar{j})}\right]\nonumber
  \end{eqnarray}
  $\bar{j}$ means that term does not exist if $j=m$.

  In addition, to calculate $\left(
        \begin{array}{cc}
          \mathrm{diag}(\mathbf{r}) & \overline{M^{(\lambda)}}\\
          \overline{M^{(\lambda)}}^\top & \mathrm{diag}(\mathbf{\bar{c}})\\
        \end{array}
      \right)^{-1}$, we can use the formula $$\left(
    \begin{array}{cc}
      A&B\\
      C&D\\
    \end{array}
  \right)^{-1}=\left(
    \begin{array}{cc}
      M & -MBD^{-1}\\
      -D^{-1}CM & D^{-1}+D^{-1}CMBD^{-1}\\
    \end{array}
  \right)$$
  where $M=(A-BD^{-1}C)^{-1}$.

  For $\rho$:
  \begin{equation}
    \dfrac{\partial\rho}{\partial M_{ij}}=\lambda M_{ij}^{\lambda-1}E(i,j)
  \end{equation}
  with $E(i,j)$ a $n\times m$-matrix where $E(i,j)_{ij}=1$ and all other entries
  vanish. And
  \begin{equation}
    \nabla_{(\mathbf{r},\mathbf{c})}\rho = \mathbf{0}.
  \end{equation}
  
  For $\mu$:
  \begin{equation}
    \dfrac{\partial \mu}{\partial\alpha_i}(P,\alpha,\beta)=
    \lambda\mathrm{diag}(\delta_i\lambda\alpha) P \mathrm{diag}(\lambda\beta)=
    \lambda P^\ast_{(i,\_)}\nonumber
  \end{equation}
  where $P^\ast_{(i,\_)}$ is a matrix with $i$-th row the same as $i$-th row of $P^\ast$ and
  vanishes elsewhere.

  Similarly,
  \begin{equation}
    \dfrac{\partial \mu}{\partial\beta_j}(P,\alpha,\beta)=
    \lambda\mathrm{diag}(\lambda\alpha)P\mathrm{diag}(\delta_j\lambda\beta)=
    \lambda P^\ast_{(\_,j)}\nonumber
  \end{equation}
  with $j\le m-1$ but the size of $P^\ast_{(\_,j)}$ is still $n\times m$.

  And
  \begin{equation}
    \dfrac{\partial\mu}{\partial P_{ij}} = \mathrm{diag}(\lambda\alpha)E(i,j)\mathrm{diag}(\lambda\beta)
    =\dfrac{P^\ast_{ij}}{P_{ij}}E(i,j)\nonumber
  \end{equation}
  where $P^\ast$ is the $(\mathbf{r},\mathbf{c})$-Sinkhorn scaling limit matrix of $P$.

  Finally, we can combine all the results above to calculate the gradient of
  $\Phi$. We will use $(\alpha,\beta)$ for $\Psi$, use $P$ for $\rho$ when it is
  convenient.

  \begin{eqnarray}
    \left(\nabla_{\mathbf{r}}\Phi\right)_t
    &=&\dfrac{\partial\Phi}{\partial\mathbf{r}_t}\nonumber\\
    &=&\sum_{i,j=1}^{n,m}\dfrac{\partial\mu}{\partial\rho_{ij}}\dfrac{\partial\rho_{ij}}{\partial\mathbf{r}_t}
        +\sum_{i=1}^{n}\dfrac{\partial\mu}{\partial\alpha_i}\dfrac{\partial\alpha_i}{\partial\mathbf{r}_t}
        +\sum_{j=1}^{m-1}\dfrac{\partial\mu}{\partial\beta_j}\dfrac{\partial\beta_j}{\partial\mathbf{r}_t}
        \nonumber\\
    &=&0+\sum_{i=1}^{n}\left(\dfrac{\partial\Psi_i}{\partial\mathbf{r}_t}\right)\dfrac{\partial\mu}{\partial\alpha_i}+
      \sum_{j=1}^{m-1}\left(\dfrac{\partial\Psi_{n+j}}{\partial\mathbf{r}_t}\right)\dfrac{\partial\mu}{\partial\beta_j}
        \nonumber
  \end{eqnarray}

  If we write the column $t$ of matrix $\left(
        \begin{array}{cc}
          \mathrm{diag}(\mathbf{r}) & \overline{M^{(\lambda)}}\\
          \overline{M^{(\lambda)}}^\top & \mathrm{diag}(\bar{\mathbf{c}}))\\
        \end{array}
      \right)^{-1}$
      in terms of $\left(
        \begin{array}{c}
          \mathbf{u}\\
          \bar{\mathbf{v}}\\
        \end{array}
      \right)$ with $\mathbf{u}\in\mathbb{R}^n$ and
      $\mathbf{v}\in\mathbb{R}^{m}$
      with the last entry $\mathbf{v}_m=0$ then
  
  \begin{eqnarray}
    \left(\nabla_{\mathbf{r}}\Phi\right)_t
    &=&-\mathrm{diag}\left(\mathbf{u}\right)M^{(\lambda)}-M^{(\lambda)}\mathrm{diag}\left(\mathbf{v}\right)\nonumber
  \end{eqnarray}
  
  To calculate $\nabla_{\mathbf{c}}\Phi$, we choose
  an elegant way by using the above calculations.
  We rewrite the map $\Phi$ as
  $\Phi(M,\mathbf{r},\mathbf{c})=(\Phi(M^\vee,\mathbf{r}^\vee,\mathbf{c}^\vee))^\top$ with 
  $M^\vee=M^\top$, $\mathbf{r}^\vee=\mathbf{c}$ 
  and $\mathbf{c}^\vee=\mathbf{r}$. The transpose of $M$, after regularization, scaled to $(\mathbf{c},\mathbf{r})$ is exactly $(M^{(\lambda)})^\top$.
  
  So we have
  $\nabla_{\mathbf{c}}\Phi(M,\mathbf{r},\mathbf{c})=\nabla_{\mathbf{r}^\vee}(\Phi(M^\vee,\mathbf{r}^\vee,\mathbf{c}^\vee))^\top$, thus
  \begin{eqnarray}
    (\nabla_{\mathbf{c}}\Phi(M,\mathbf{r},\mathbf{c}))_s
    &=&((\nabla_{\mathbf{r}^\vee}\Phi(M^\vee,\mathbf{r}^\vee,\mathbf{c}^\vee))_s)^\top\nonumber\\
    &=&-M^{(\lambda)}\mathrm{diag}(\mathbf{u})-\mathrm{diag}(\mathbf{v})M^{(\lambda)},\nonumber
  \end{eqnarray}
  where $\left(\begin{array}{c}
       \mathbf{u}\\
       \bar{\mathbf{v}}
  \end{array}\right)$
  is the $s$-th column of matrix 
  $\left(\begin{array}{cc}
       \mathrm{diag}(\mathbf{c})& \overline{M^{(\lambda)}}^\top \\
       \overline{M^{(\lambda)}}& \mathrm{diag}(\bar{\mathbf{r}})
  \end{array}\right)^{-1}.$
  

  At last, 
  \begin{eqnarray}
    \left(\nabla_{M}\Phi\right)_{st}
    &=&\dfrac{\partial\Phi}{\partial M_{st}}\nonumber\\
    &=&\sum_{i,j=1}^{n,m}\dfrac{\partial\mu}{\partial\rho_{ij}}\dfrac{\partial\rho_{ij}}{\partial
        M_{st}}
        +\sum_{i=1}^{n}\dfrac{\partial\mu}{\partial\alpha_i}\dfrac{\partial\alpha_i}{\partial
        M_{st}}
        +\sum_{j=1}^{m-1}\dfrac{\partial\mu}{\partial\beta_j}\dfrac{\partial\beta_j}{\partial
        M_{st}}
        \nonumber\\
    &=&\lambda\dfrac{M^{(\lambda)}_{st}}{M_{st}} \left( E(s,t)-
        \mathrm{diag}\left(\mathbf{u}\right)M^{(\lambda)}-M^{(\lambda)}\mathrm{diag}\left(\mathbf{v}\right)
        \right)\nonumber
  \end{eqnarray}
  where $\mathbf{u}\in\mathbb{R}^n$, $\mathbf{v}\in\mathbb{R}^{m}$ with the
  last entry $\mathbf{v}_m=0$, and $$\left(
    \begin{array}{c}
      \mathbf{u}\\
      \bar{\mathbf{v}}\\
    \end{array}
  \right)=\left[\left(
        \begin{array}{cc}
          \mathrm{diag}(\mathbf{r}) & \overline{M^{(\lambda)}}\\
          \overline{M^{(\lambda)}}^\top & \mathrm{diag}(\bar{\mathbf{c}}))\\
        \end{array}
      \right)^{-1}_{\text{col-}s}+\left(
        \begin{array}{cc}
          \mathrm{diag}(\mathbf{r}) & \overline{M^{(\lambda)}}\\
          \overline{M^{(\lambda)}}^\top & \mathrm{diag}(\bar{\mathbf{c}}))\\
        \end{array}
      \right)^{-1}_{\text{col-}(n+\bar{t})}\right],$$
  for $(s,t)\in\mathfrak{P}$, and $\bar{t}$ means that term does not exist if $t=m$.

\end{proof}

\end{document}